%% file: paper.tex
\documentclass{article}

\usepackage[dblblindworkshop,final]{neurips_2026}
\renewcommand{\acksection}{\section*{Acknowledgments}}
\workshoptitle{Managing Agents that Manage Agents:
  Workshop on Responsible Use of Meta-Agents}

\usepackage[utf8]{inputenc}
\usepackage[T1]{fontenc}
\usepackage{amsmath,amssymb}
\usepackage{booktabs}
\usepackage{graphicx}
\usepackage{microtype}
\usepackage[table]{xcolor}
\usepackage{fontawesome5}
\usepackage{hyperref}
\usepackage{comment}
\definecolor{trophygold}{HTML}{B58900}
\newcommand{\trophy}{\textcolor{trophygold}{\faTrophy}}
\newcommand{\runlink}[2]{\href{https://github.com/andborth/RoboPhD/tree/main/example_runs/robophd/#1/#2}{\scriptsize\faExternalLink*}}
\newcommand{\agentlink}[3]{\href{https://github.com/andborth/RoboPhD/blob/main/example_runs/robophd/#1/#2/agents/#3/agent.py}{\scriptsize\faExternalLink*}}
\hypersetup{colorlinks=true,linkcolor=blue!60!black,citecolor=blue!60!black,urlcolor=blue!60!black}

\newcommand{\robophd}{RoboPhD}
\newcommand{\snapdate}{2026-11-20}
\newcommand{\dsk}{DS-1000}
\newcommand{\pfb}{PaperFindingBench}

\title{Competing at Every Price Point with\\Agentic Evolution over a Menu of LLMs}

\author{%
  Andrew Borthwick\\
  Independent Researcher\\
  \texttt{aeborthwick@gmail.com}\\
}

\begin{document}

\maketitle

\begin{abstract}
Consider a firm that surveys its competition for a particular agentic task
and seeks to offer superior accuracy at every price point.
A firm that Pareto-dominated its competitors would leave no rational
customer a reason to buy elsewhere. This paper shows a path to this kind
of capability by evolving multi-LLM Python agents from training
pools of at most 100 examples.
Given a priced menu of nine LLM endpoints; brief documentation of the task, objective,
and API; a simple seed agent; and an
operator-chosen per-problem cost target---usually set at an incumbent's own price---\robophd{},
an evolutionary meta-agent, evolves complete agent programs that attack the public frontiers of
two semantically dissimilar tasks point by point: \dsk{} (execution-checked code generation)
and \pfb{} (LLM-judged scientific document retrieval). On public leaderboards for each task, the evolved agents hold every
Pareto-frontier slot on \pfb{} and all but one on \dsk{}, including Pareto domination of both
the top-scoring and the lowest-cost competing points.
\end{abstract}

\section{Introduction}
\label{sec:intro}

Industrial deployment of an LLM agent almost inevitably involves a decision about the tradeoff between quality and inference cost. It is frequently possible to gain higher levels of accuracy by spending more on inference, but every customer is likely to have a different budget and a different tolerance for error. Thus, to maximize sales, the vendor of an agentic solution for a task would seek to Pareto-dominate every competitor's offering on the price and quality axes.  

Yet current tools for \emph{building} agents optimize almost
exclusively the quality axis. The recent wave of harness-optimization
systems---Meta-Harness \citep{lee2026metaharness}, HARBOR
\citep{sengupta2026harbor}, Self-Harness \citep{zhang2026selfharness},
HarnessForge \citep{chen2026harnessforge}---searches over harness code
or configuration around a \emph{fixed} model, reporting cost (when at
all) as an outcome rather than a target. Meanwhile a team that needs an
agent at a specific price point has one crude lever today: swap the
model inside a fixed scaffold. Delivering the best value at a particular price point thus becomes a hit-and-miss affair.  This approach also inhibits the discovery of more sophisticated cost-saving strategies, such as using a cheaper model for routine steps and reserving a stronger model for hard cases or adjudication.

\begin{figure}[t]
  \centering
  \includegraphics[width=\linewidth]{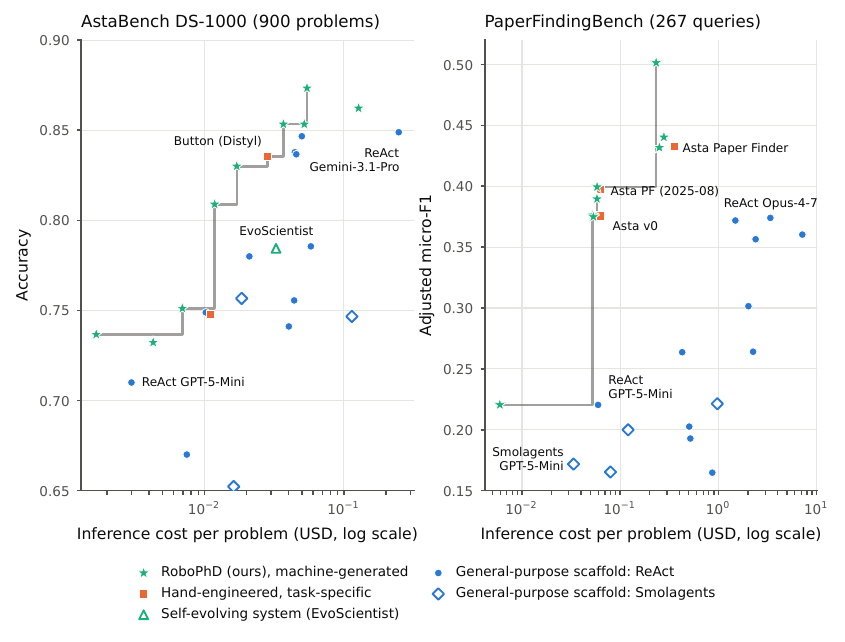}
  \caption{Cost--quality frontiers, AstaBench \dsk{} (left) and \pfb{}
  (right), leaderboard snapshot \snapdate{}. The gray staircase is each board's Pareto
  frontier: \robophd{} holds six of its seven points on \dsk{} (all but
  the hand-built Button) and all five on \pfb{}. The full boards are
  Tables~\ref{tab:dsboard}--\ref{tab:pfboard}.
  }
  \label{fig:money}
\end{figure}

We report here a demonstration of this capability on two public leaderboards\footnote{See \pfb{} and \dsk{} at \url{https://allenai-asta-bench-leaderboard.hf.space/}.}
from AstaBench \citep{bragg2026astabench} (Figure~\ref{fig:money}),
achieved by evolving complete agent Python programs 300--2{,}800 lines long, each incorporating calls to between one and six models from a priced menu of nine LLM endpoints. We chose the two AstaBench tasks to be
semantically dissimilar while each proxying a major industrial workload
(Table~\ref{tab:arenas}):
\dsk{} \citep{lai2023ds1000} is data-science \emph{code generation}---
self-evidently commercial---while \pfb{} is ranked \emph{document
retrieval} with cited evidence over a massive scientific corpus, accessed
through a provided tool suite. The latter task is both directly useful to
scientists (one of the competing entries is a deployed system) and
analogous to the common industrial problem of searching a large
organization's internal documents. The pair also
covers two
distinct evaluation regimes: \dsk{} is execution-checked against
gold-standard tests, while 73\% of \pfb{} queries are scored by an
LLM judge over agent-supplied evidence. 

This paper exercises the \robophd{} engine\footnote{Code:
\url{https://github.com/andborth/RoboPhD}.} of \citet{robophd2026engine},
where its algorithm, diversity mechanisms, and a controlled four-task
engine comparison are documented. Contributions of this work include: \textbf{(1)}~cost-\emph{targeted}
agent evolution: the operator names a per-query dollar target, and the
training score is unpenalized below it and linearly penalized above it,
a design intended to make the budget a resource to spend rather than a
cost to minimize (Section~\ref{sec:system}); \textbf{(2)}~evolution over
a priced multi-provider model menu, with the choice of model at each call
site in the evolved artifact left to the discretion of the evolutionary meta-agent, so that the combined evolutionary pressure of accuracy and cost
is answered by \emph{discovered composition} (cascades,
ensembles, stage-wise model and effort assignment, spanning one to six
models across our entries)
(Section~\ref{sec:system}); \textbf{(3)}~two-task results showing domination of the entire \pfb{} Pareto frontier and all but one point of the \dsk{} frontier against
public, independently scored leaderboards, achieved under data-starved training pools
of 66 and 100 examples
(Sections~\ref{sec:ds1000}--\ref{sec:pfb}); \textbf{(4)}~a management taxonomy with field
reports on incentive design for an evolutionary optimizer
(Section~\ref{sec:manage}).

\section{System summary}
\label{sec:system}

\paragraph{\robophd{} in a nutshell.}
A run keeps a pool of candidate agents---each a single Python
file---and repeats an iterative cycle until its evaluation budget is spent. Each
iteration draws a fresh random batch of training examples (14 on
\pfb{}, 20 on \dsk{}) and scores three agents on it: the previous
iteration's winner, a newly evolved agent, and one drawn at random from the
next two on the Elo table. Elo rankings are updated from pairwise 
win/loss comparisons of agent mean scores on an iteration's batch of training examples.
An ``evolution session''---the evolution model running in Claude Code with
file and shell access to the run's workspace---then reads the
competitors' code, per-example diagnostics, and a report of where they
diverged, and writes the next candidate. The run ships its final Elo
leader. Algorithm details, diversity mechanisms, and a
controlled four-task comparison against GEPA \citep{gepa2025} and
Karpathy's Autoresearch \citep{autoresearch2026} under matched budgets and seed agents are in \citet{robophd2026engine}. Here we show the
engine generalizes to two unrelated data-starved task families and add
two capabilities: cost targeting and evolving over a priced
multi-provider menu of LLMs.

\paragraph{A novel average cost objective.}

Prior work \citep{gepa_oa2026,robophd2026engine} has enforced cost constraints
example by example, penalizing the optimizer for exceeding a fixed
budget on any single one. However, the metric that matters---on the AstaBench boards we are targeting as well as in ordinary
business budgeting---is \emph{average} cost. An operator typically accepts that some
queries may cost more if others cost less. 
\robophd{}'s core structure of comparing the scores of three agents over batches of 
14--20 examples naturally accommodates a training objective which directly 
combines average cost with average score.
For each training batch, we combine aggregated
per-example cost and quality as follows:
\begin{equation}
\label{eq:objective}
S \;=\; \frac{100}{n}\,\Biggl[\;\underbrace{\sum_i q_i}_{\text{correct answers}}
\;-\;
\underbrace{\max\!\Bigl(0,\; \frac{\bar{c} - \tau}{\lambda}\Bigr)}_{\text{cost penalty, in error-equivalents}}\Biggr]
\end{equation}
where $n$ is the batch size, $q_i \in [0,1]$ is example $i$'s quality
score (binary on \dsk{}, mostly F1 on \pfb{}), $\bar{c}$ is the batch's
mean per-example cost,\footnote{Per AstaBench convention, only agent LLM usage is metered. LLM judge usage in
\pfb{} is considered free.} $\tau$ is the operator's cost threshold, and
$\lambda$, the \emph{overage per error}, is the mean per-query overspend,
in dollars, that costs as much as one fully wrong query; a smaller
$\lambda$ is thus a steeper penalty (Table~\ref{tab:penalty} in
Appendix~\ref{app:inputs} shows how we describe this to the evolutionary
meta-agent). The penalty exists only at training time, as the
signal steering evolution toward the requested operating point; at test time, 
we report raw score and cost as separate axes,
so no number in Sections~\ref{sec:ds1000}--\ref{sec:pfb} is
penalty-adjusted. 
The linear ramp makes a small breach a small matter,
letting evolution work near the boundary of the ``free zone,'' $\tau$. The overage per error
$\lambda$ is itself settable---an incentive-design knob whose effect we
examine in Section~\ref{sec:manage},
item~3---and our later runs default to
$\lambda = 0.1\tau$, scaling the penalty with the target: exceeding a
budget by \$0.01 should matter far more against a \$0.02 target than
against a \$0.20 one.

Note that in this work we think of cost as a ``free''
resource to be consumed below a run's cost target of $\tau$.  Out of a 
concern that this might contradict an LLM's typical prior of treating 
frugality as a virtue in itself, we include this in the opening
instruction of its objective document:
 \emph{``your primary goal is simple: maximize
the score on held-out queries. The scoring function [\dots] encodes this
directly---retrieval quality is the dominant signal, and cost acts as a
tiebreaker close to threshold but starts to actively trade off against
score farther out.''}  An indication that this objective was understood
is that the meta-agent behind the \dsk{} \$0.037 entry, measuring its
early agents at $\sim$1.5\% of the \$0.08 threshold, recorded in its
reasoning: \emph{``Cost is not a constraint; accuracy is the only
lever. I have $\sim$60$\times$ headroom to spend on stronger models,
reasoning, and multi-sample verification before the cost penalty even
begins.''}

\paragraph{Evolving agents over a menu of models.}
So as to give the evolutionary process more freedom to achieve its accuracy
goals within its cost budget, we propose agentic evolution over a \emph{menu} of 
LLMs.  We provide evolution with a registry of nine priced handles---three
each from OpenAI, Anthropic, and Google, spanning roughly $10\times$ in
per-token price, each with a per-call \texttt{reasoning\_effort} knob
(Appendix~\ref{app:tasks}) and give evolution complete discretion on how to 
exploit these options.
On both tasks, we start from a seed agent that calls only a single model, GPT-5.4-Mini, but the joint pressures of cost and accuracy drive the evolving agents to different kinds of composition---cheap-first cascades with strong-model escalation,
mixed ensembles with adjudication, model choice per pipeline stage---rather
than frugality within one model. The range this opens is wide and is
used: the seventeen board entry agents use between one and six models. The top \dsk{} entry (0.873 at
\$0.054) runs a three-provider ensemble and escalates to five handles only when its candidates disagree; the 0.809 entry
delivers near-bottom cost (\$0.012/problem) on a comparatively expensive
model, Claude-Sonnet-4.6, spent sparingly, while the \$0.004 entry
composes the three cheapest-tier models plus a sliver of GPT-5.4. Collectively the seventeen entries exercise all nine handles on the menu.

\paragraph{A system with minimal human inputs.}
Of \robophd{}'s inputs, the dataset and evaluator are functionally free---we contend that any
problem worth solving already has data and a way to score an attempt (a
slogan for \robophd{}: ``if you can benchmark it, \robophd{} can optimize
it''). The
authored increment is small: a five-line objective; a background document (130 to 227 lines) describing the
task---scoring rules, output schema, and the tool and model
APIs---largely transcribed from existing documentation and, by house
rule, containing no strategy hints or recommended settings; and a minimal seed agent (46
to 121 lines) which makes one or two calls per problem to a single cheap
model, GPT-5.4-Mini. The objectives are reproduced in full, and the background
documents in verbatim excerpt, in Appendix~\ref{app:inputs}.
From this, evolution produced frontier-holding agents of 302--2,781
lines---every cascade, ensemble, and model choice in them discovered. In addition to these modest human inputs, a \robophd{} evolutionary run incurs \$34--\$368 of compute. By contrast, the announcement for Ai2's Asta Paper Finder credits a
fourteen-person team \citep{asta_paper_finder}.

\paragraph{Three classes of competitor.}
Both boards are populated by two classes, and our entries constitute a third:
\textbf{(1)}~general-purpose scaffolds swept over models (ReAct
\citep{yao2023react}, Smolagents Coder \citep{smolagents}, submitted by the
board maintainers)
---zero task engineering, sparse cost points, the model swap being the only
cost dial; \textbf{(2)}~hand-engineered task-specific systems (Distyl AI's
commercial Button on \dsk{}; Ai2's Asta Paper Finder
\citep{asta_paper_finder} and Asta v0
\citep{bragg2026astabench})---expert-built, holding top slots;
\textbf{(3)}~\emph{machine-generated} task-specific agents (ours)---running
on AstaBench's Standard tooling tier.\footnote{AstaBench tags every entry with a
\emph{tooling tier}. \emph{Standard} means the agent uses only the
benchmark's canonical toolset---for these tasks, the sandboxed Python
environment and the Asta corpus API---the same tools Ai2's own ReAct
baselines run on.}

\section{\dsk{}: execution-checked code generation}
\label{sec:ds1000}

\begin{table}[t]
\centering
\small
\setlength{\tabcolsep}{5pt}
\renewcommand{\arraystretch}{1.22}
{\arrayrulecolor{gray!55}
\begin{tabular}{|l|p{1.65in}|p{2.00in}|}
\hline
 & \dsk{} & \pfb{} \\
\hline
Task & data-science code generation & ranked scientific literature retrieval \\
\hline
Verifier & hidden execution tests (no judge) & exact match (27\%) $+$ LLM judge over agent-supplied evidence (73\%) \\
\hline
Per-problem score & binary & continuous F1 in $[0, 1]$ \\
\hline
Tools (free) & Python sandbox & Asta MCP corpus API, eight tools \\
\hline
Train pool / held-out & 100 / 900 & 66 / 267 \\
\hline
Batch / evals per run & 20 / $\sim$580--800 & 14 / $\sim$600 \\
\hline
Best generic scaffold & 0.849 @ \$0.247 (ReAct/Gemini-3.1-Pro) & 0.374 @ \$3.381 (ReAct/Opus-4-7) \\
\hline
Best hand-built & 0.836 @ \$0.028 (Button) & 0.433 @ \$0.355 (Asta Paper Finder) \\
\hline
Best \robophd{} & 0.873 @ \$0.054 & 0.501 @ \$0.232 \\
\hline
Our $\tau$ range & \$0.003--\$0.16 & \$0.033--\$0.355 \\
\hline
\end{tabular}}
\arrayrulecolor{black}
\caption{The two tasks studied in this work.}
\label{tab:arenas}
\end{table}

\paragraph{Setup.}
AstaBench \dsk{} poses data-science coding problems originally
collected from Stack Overflow questions about seven Python libraries
(NumPy, Pandas, SciPy, Matplotlib, Scikit-learn, TensorFlow, PyTorch),
many perturbed against memorization \citep{lai2023ds1000}: each problem
gives a natural-language question with a partial code context, and the
agent completes the code. Scoring is binary per problem, by hidden
execution tests (some with code-style constraints); there is no judge
(See Table~\ref{tab:arenas} and 
Appendix~\ref{app:examples} for details and an example problem).

\begin{table}[t]
\centering
\footnotesize
\setlength{\tabcolsep}{3.5pt}
\begin{tabular}{@{}cllp{0.85in}lrc@{}}
\toprule
 & Entry & Score @ \$/prob & $\tau$ (\$; aimed at) & Evol.\ model & Util. & Run \\
\midrule
\multicolumn{7}{@{}l}{\dsk{}} \\
\rowcolor{gray!12} \trophy{} & RoboPhD & 0.873 @ \$0.054 & 0.127 (board top) & Fable-5.1 & 43\% & \runlink{asta_ds1000}{v0_1_0_cap_0_127_fable51} \\
\rowcolor{gray!12} & RoboPhD & 0.862 @ \$0.127 & 0.16 & Opus-4.7 & 80\% & \runlink{asta_ds1000}{v0_0_1_cap_0_16_opus} \\
\rowcolor{gray!12} & RoboPhD & 0.853 @ \$0.052 & 0.06 & Fable-5 & 87\% & \runlink{asta_ds1000}{v0_0_3_cap_0_06_fable} \\
\rowcolor{gray!12} \trophy{} & RoboPhD & 0.853 @ \$0.037 & 0.08 & Opus-4.8 & 46\% & \runlink{asta_ds1000}{v0_0_4_cap_0_08_opus} \\
 & \emph{4 entries, dominated} & 0.837--0.849 @ \$0.044--\$0.247 & \multicolumn{2}{l}{\emph{ReAct}} & --- &  \\
\trophy{} & Button & 0.836 @ \$0.028 & --- & --- & --- &  \\
\rowcolor{gray!12} \trophy{} & RoboPhD & 0.830 @ \$0.017 & 0.05 & Opus-4.8 & 34\% & \runlink{asta_ds1000}{v0_0_5_cap_0_05_opus} \\
\rowcolor{gray!12} \trophy{} & RoboPhD & 0.809 @ \$0.012 & 0.08 & Opus-4.7 & 15\% & \runlink{asta_ds1000}{v0_0_2_cap_0_08_opus} \\
 & \emph{5 entries, dominated} & 0.756--0.786 @ \$0.018--\$0.058 & \multicolumn{2}{l}{\emph{ReAct, Smolagents, EvoScientist}} & --- &  \\
\rowcolor{gray!12} \trophy{} & RoboPhD & 0.751 @ \$0.007 & 0.05 & Opus-4.8 & 14\% & \runlink{asta_ds1000}{v0_0_5_cap_0_05_opus_deep_focus} \\
 & \emph{4 entries, dominated} & 0.741--0.749 @ \$0.010--\$0.114 & \multicolumn{2}{l}{\emph{incl.\ hand-built Asta v0}} & --- &  \\
\rowcolor{gray!12} \trophy{} & RoboPhD & 0.737 @ \$0.002 & 0.003 (ReAct) & Opus-4.8 & 56\% & \runlink{asta_ds1000}{v0_0_7_cap_0_003_opus} \\
\rowcolor{gray!12} & RoboPhD & 0.732 @ \$0.004 & 0.003 (ReAct) & Fable-5 & 143\%$^{\S}$ & \runlink{asta_ds1000}{v0_0_6_cap_0_003_fable} \\
 & ReAct/GPT-5-Mini & 0.710 @ \$0.003 & --- & --- & --- &  \\
 & \emph{11 entries, dominated} & 0.027--0.670 @ \$0.004--\$0.137 & \multicolumn{2}{l}{\emph{ReAct, Smolagents}} & --- &  \\
\midrule
\multicolumn{7}{@{}l}{\pfb{}} \\
\rowcolor{gray!12} \trophy{} & RoboPhD & 0.501 @ \$0.232 & 0.279 (board top) & Fable-5.2 & 83\% & \runlink{asta_paper_finder}{v0_1_1_cap_0_279_fable52} \\
\rowcolor{gray!12} & RoboPhD & 0.440 @ \$0.279 & 0.355 (Asta PF) & Fable-5 & 78\% & \runlink{asta_paper_finder}{v0_0_9_cap_0_355_fable} \\
 & Asta Paper Finder & 0.433 @ \$0.355 & --- & --- & --- &  \\
\rowcolor{gray!12} & RoboPhD & 0.432 @ \$0.251 & 0.355 (Asta PF) & Opus-5 & 71\% & \runlink{asta_paper_finder}{v0_0_9_cap_0_355_opus5} \\
\rowcolor{gray!12} \trophy{} & RoboPhD & 0.399 @ \$0.058 & 0.063 (Asta PF) & Fable-5.1 & 92\% & \runlink{asta_paper_finder}{v0_1_0_cap_0_063_fable51} \\
 & Asta Paper Finder & 0.397 @ \$0.063 & --- & --- & --- &  \\
\rowcolor{gray!12} \trophy{} & RoboPhD & 0.390 @ \$0.058 & 0.063 (Asta PF) & Fable-5 & 92\% & \runlink{asta_paper_finder}{v0_0_9_cap_0_063_fable} \\
\rowcolor{gray!12} \trophy{} & RoboPhD & 0.376 @ \$0.052 & 0.063 (Asta PF) & Opus-5 & 83\% & \runlink{asta_paper_finder}{v0_0_9_cap_0_063_opus5} \\
 & Asta v0 & 0.376 @ \$0.063 & --- & --- & --- &  \\
\rowcolor{gray!12} & RoboPhD & 0.375 @ \$0.053 & 0.06 (ReAct) & Fable-5 & 89\% & \runlink{asta_paper_finder}{v0_0_7_cap_0_06_fable} \\
 & \emph{8 entries, dominated} & 0.221--0.374 @ \$0.43--\$7.16 & \multicolumn{2}{l}{\emph{ReAct, Smolagents}} & --- &  \\
\rowcolor{gray!12} \trophy{} & RoboPhD & 0.220 @ \$0.006 & 0.033 (Smolagents) & Opus-4.8 & 18\% & \runlink{asta_paper_finder}{v0_0_8_cap_0_033_opus} \\
 & ReAct/GPT-5-Mini & 0.220 @ \$0.060 & --- & --- & --- &  \\
 & \emph{3 entries, dominated} & 0.193--0.203 @ \$0.12--\$0.52 & \multicolumn{2}{l}{\emph{ReAct, Smolagents}} & --- &  \\
 & Smolagents/GPT-5-Mini & 0.172 @ \$0.033 & --- & --- & --- &  \\
 & \emph{10 entries, dominated} & 0.046--0.165 @ \$0.01--\$2.82 & \multicolumn{2}{l}{\emph{ReAct, Smolagents}} & --- &  \\
\bottomrule
\end{tabular}
\caption{All seventeen \robophd{} submissions (shaded), in board order,
interleaved with the competitor field. \trophy{}: on the board's Pareto frontier (official
semantics); un-trophied \robophd{} rows are displaced by their cheaper
neighbors. $\tau$ provenance in
parentheses where set at another entry's price. $^{\S}$: official-basis
utilization, partly measurement error. Run: link to the canonical snapshot of the evolutionary run, including
the shipped agent with its full lineage, key reports, and the
meta-agent's reasoning and reflection.
Full boards:
Tables~\ref{tab:dsboard}--\ref{tab:pfboard}; run detail:
Table~\ref{tab:runs}.}
\label{tab:submissions}
\end{table}

\paragraph{Frontier results.}
We submitted nine evolved \robophd{} agents to be scored by Ai2's official \texttt{astabench}
evaluation (Table~\ref{tab:submissions}; Figure~\ref{fig:money}). The net result:
every ReAct and Smolagents entry on the \dsk{} board is Pareto-dominated
by a \robophd{} point, and the only non-\robophd{} entry left on the
frontier is the hand-engineered Button.  On the other hand, note that
Ai2's hand-built, fully-custom Asta v0 is dominated by one of our
cheapest entries, and the board's nearest neighbor to our class---
EvoScientist-Code, the coding arm of a self-evolving multi-agent ``AI
scientist'' \citep{lyu2026evoscientist}---is dominated as well.

\paragraph{What evolution built.}
The frontier is held by qualitatively different agents (see Appendix~\ref{app:arch} for details). The \$0.054 agent (1,226 lines~\agentlink{asta_ds1000}{v0_1_0_cap_0_127_fable51}{iter6_ensemble_exec_vote_escalate}) has three models from the three
providers write candidates, executes each in the sandbox, and accepts
an execution-output majority without any judge call; only on
disagreement does it add voices or an adjudicating judge, and it
safely rewrites looping answers into the vectorized form that \dsk{}'s
hidden style rules sometimes demand.
The \$0.037 agent (500 lines~\agentlink{asta_ds1000}{v0_0_4_cap_0_08_opus}{iter11_ds1000_tridtype_judge}) executes three medium-cost models from
different model families at baseline effort settings, compares outputs
down to DataFrame and array dtypes, and escalates to a high-effort model
for adjudicating disagreements.
Evolution made an intriguing choice on the \$0.002 agent~\agentlink{asta_ds1000}{v0_0_7_cap_0_003_opus}{iter12_strong_toplevel}.  Rather than rely on the lowest-cost OpenAI model, GPT-5.4-mini, it went with a one-shot call to the stronger GPT-5.4, a 59-word preamble, and some deterministic format repair---only 302 lines of Python code.  It tried and rejected (under this budget) heavy prescriptive preambles and the sandbox execution of candidate solutions of the kind used by the \$0.037 agent.

\section{\pfb{}: LLM-judged scientific document retrieval}
\label{sec:pfb}

\paragraph{Setup.}
\pfb{} poses natural-language literature-search queries in three
classes: 38 \texttt{specific} (one paper the user can already name---``the
original transformers paper''), 35 \texttt{metadata} (a set fixed by
bibliographic constraint---``2012 papers by David Harel''), and 194
\texttt{semantic} (a topical need with no closed gold list---``papers on
online adaptation of neural MT metrics at inference'').  See 
Appendix~\ref{app:examples} for additional examples. 
The two exact-match classes ($\sim$27\%) are scored by ID intersection
against an enumerated gold set.  For the semantic majority, the agent
returns each paper's ID with a set of snippets drawn from that paper, and
an LLM judge adjudicates the paper's relevance from the snippets alone.
In all three query classes, the agent is provided with eight tools over the
Semantic Scholar corpus \citep{kinney2023semanticscholar}: keyword, title, full-text-snippet, and author search, plus paper metadata, batch lookup, and citation traversal.  See 
Appendix~\ref{app:inputs} for API details.

\paragraph{Frontier results.}
As evaluated by Ai2, our eight
submitted agents Pareto-dominate the entire \pfb{} leaderboard
(Table~\ref{tab:submissions}; Figure~\ref{fig:money}). At the top, our agent with a result of
0.501 at \$0.232 beats the best entry from any other team, Ai2's
hand-engineered Asta Paper Finder (0.433 at \$0.355), by seven percentage points at 35\% lower cost. 
At the cheap end,
our 0.220 at \$0.006 agent matches ReAct/GPT-5-Mini's score at a tenth of its
cost.

\paragraph{What evolution built (lower-cost agents).}
The \$0.053 agent (2,097 lines~\agentlink{asta_paper_finder}{v0_0_7_cap_0_06_fable}{iter12_body_conjunction}) plans with GPT-5.4, retrieves by
\emph{body-text conjunction}---conjunctive corpus queries over full-text
matches---and grades candidates with a GPT-5.4-mini cascade before
assembling grounded evidence. The \$0.006 agent (1,006 lines~\agentlink{asta_paper_finder}{v0_0_8_cap_0_033_opus}{iter9_rerank_rich_v1}) makes just
two calls per query, both to GPT-5.4-mini: one to turn the request into
searches and one to rank what they return. All the work in between is
done by search-tool calls, which are free. It matches the score of the
ReAct/GPT-5-Mini baseline at a tenth of its cost.

\paragraph{The board leader and a model-menu ablation.}
The \$0.232 board leader (1,811 lines~\agentlink{asta_paper_finder}{v0_1_1_cap_0_279_fable52}{iter20_year_steered_expansion}, six models from all three
providers) divides its work by price. A Claude-Opus-4.8 planner sits in
front of every query: because the benchmark's criteria for judging a
paper's relevance are hidden, Opus infers what they should be, along
with their relative weights. GPT-5.4-mini then grades every
candidate against each criterion, the agent follows citations in both
directions from the papers that satisfy them all, and GPT-5.4 re-judges
the strongest (Appendix~\ref{app:arch}). 

Prior evolutionary approaches
\citep{robophd2026engine,gepa_oa2026} evolved agents around a single,
predetermined model, so this division of labor is something they could
not have built. To test whether it matters, we repeated the leader's
evolutionary run twice at the same cap ($\tau = \$0.279$) with the menu
restricted to one model, GPT-5.4-mini in one arm and GPT-5.4 in the
other. A single-model agent must buy planning, grading volume, and final
judgment at one price, and the two arms gave up opposite things. The
GPT-5.4-mini agent keeps the volume, grading up to 930 candidates per
query, but plans and re-judges with the same small model. The GPT-5.4
agent has the stronger model throughout but grades at most 212. Both
fall short of the leader on the 267 held-out queries (on a luna-judged basis, see
Appendix~\ref{app:tasks}): 0.493 and 0.459 against 0.510
(Table~\ref{tab:menuablation}). Note also that neither
arm used as much of the cap as the board leader (73\% against 83\%).
This is to be expected because single-model menus decrease the
flexibility to fully utilize a cost target. For instance, here the gap
arises on specific-paper and metadata queries, where the leader calls
its most expensive models.

\section{Coping with data starvation}
\label{sec:overfit}

\paragraph{A data-starved regime.}
Both tasks are data-starved: training pools of 66 (\pfb{}) and 100 (\dsk{})
examples against held-out test sets of 267 and 900---pools well below
the 400-example minimum training pool of the four tasks described in \citet{robophd2026engine}. However, this is also the regime many deployments actually inhabit:
labeled examples,
or even examples crisp enough for an LLM judge to score, are commonly
the scarcest input a team has. At these volumes,
weight-based adaptation---supervised or parameter-efficient fine-tuning,
RL---is impractical even setting aside its other obstacles here (the
strongest solver models are closed API endpoints, and the tasks provide
ways to \emph{score} an attempt, not demonstrations of tool-using
solution paths to imitate). Optimizing the agent \emph{program}, with an LLM
reflecting on rich per-example diagnostics, is the approach that operates
at this scale \citep{gepa_oa2026,robophd2026engine}.
Working with those larger pools, \citet{robophd2026engine} used a fixed
batch of 20 examples per iteration. Here, to minimize overlap between batches, we set the batch to $\lceil 20\%\rceil$ of the
pool: 14 for \pfb{} and 20 for \dsk{}. Submitted runs lasted 21--23
iterations on \pfb{} and 14--20 on \dsk{}, so a run sees each pool
example roughly three to five times.

\paragraph{The objective is a document, not a scalar.}
Pool overfitting is the default outcome for \emph{any} optimizer here.  The core \robophd{} algorithm of randomly drawing a fresh batch with every evolutionary iteration and Elo ranking the evolved agents described in \citet{robophd2026engine} is only a partial solution at the extreme levels of data starvation we are studying here.  The resulting evolutionary pressure would prove inadequate when it is possible to memorize the entire training pool.  Although the deployed agent only receives the query, evolution is shown full per-example ground truth.  For instance, in \pfb{} it is shown relevance criteria, per-paper judge
verdicts, and score decomposition. To meet this challenge, note that although one cannot tell SGD that the
batch is a proxy, an LLM optimizer \emph{can} be told. Both tasks'
objective documents (Appendix~\ref{app:inputs}) say so three times.  For instance, the
goal is defined as score on \emph{held-out} problems and the closing line restates the objective as \emph{``build an
agent that generalizes to unseen problems---the visible batch is a
training signal, not the target''}. 

\begin{figure}[t]
\centering
\includegraphics[width=\linewidth]{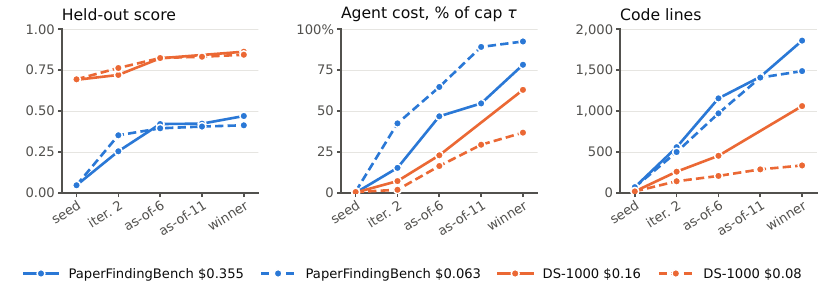}
\caption{Held-out score, agent cost as a share of the cap $\tau$, and
code size at each evolutionary waypoint, for four evolutionary runs. Every step rises on every track.  \pfb{} scores are luna-judged (Appendix~\ref{app:tasks}).  See Table~\ref{tab:waypoints} for details.}
\label{fig:waypoints}
\end{figure}

\paragraph{Experimental evidence on resistance to overfitting.}
The signature of overfitting would be training-time selection climbing
while held-out score falls. We measured four completed runs---two per
task, at each task's most expensive cost target and a mid-range one---at five waypoints each: the seed, the iteration-2 agent, the Elo leader as of iterations
6 and 11, and the shipped winner by Elo at the end of the evolutionary run. Note that the iteration-2 agent isolates the effect of simply using Claude Code with the provided information, absent any impact of evolutionary pressure.  Iterations 6 and 11 illustrate the compounding effects of further iterations of evolutionary refinement and evolutionary pressure.  
In every case, we saw monotonic score increases on the held-out test data (Figure~\ref{fig:waypoints}).  Interestingly, we also see monotonic increases in per-example cost, representing evolution incrementally increasing its exploitation of the available headroom under the cost cap, $\tau$; similarly, we see monotonic increases in lines of code.  Finally, an examination of the evolutionary meta-agent's reasoning at the evolutionary waypoints shows that it is mindful of the generalization requirement.  See these sample quotes (and additional examples in Table~\ref{tab:mechanisms}, Appendix~\ref{app:waypoints}):
\begin{itemize}
\item \emph{``Timeout-risk also generalizes to the held-out set (cold snippet
  queries are query-dependent [\ldots]), so the
  fix is not batch-overfit.''} (\pfb{} \$0.063, iteration 10)
\item \emph{``The two concrete edits [\ldots] target \textbf{structural} weak
  spots (subtle-correctness judging; unverifiable matplotlib), not the
  specific 20 problems. I deliberately do \textbf{not} try to
  special-case 284's dtype quirk [\ldots].''} (\dsk{} \$0.08, iteration 8)
\end{itemize}

\section{Managing the managing agent}
\label{sec:manage}

An appendix of \citet{robophd2026engine} reported an incident in which an evolutionary meta-agent discovered an API weakness which revealed hidden future datapoints on a forecasting task.  This weakness was avoided on future runs by more careful engineering. Analogously, early in the work reported here, we observed a different ``exploit'' in which the evolutionary meta-agent examined the results of prior evolutionary runs to gain insights into its own strategies, a violation of our policy of keeping each run self-contained.  Again, there was an engineering response which confined evolution to only viewing current run data.  We focus here, however, on a more subtle class of management issues: in some cases we had to provide additional tools and information to the evolutionary meta-agent; in others, we restricted its ability to consume unpriced resources that it was rationally exploiting in pursuit of its stated objective \citep[cf.\ the specification-gaming and reward-hacking literature:][]{amodei2016concrete,skalse2022defining,pan2022effects,krakovna2020specification}.

\paragraph{1. Proactive reward-channel hardening.}
The \pfb{} judge grades solely agent-supplied evidence: it never fetches
papers and never checks that a quoted passage exists.  
One could therefore imagine a meta-agent that built a lineage of agents which evolved increasingly loose paraphrasings of sections of candidate papers to support the claim that the paper was relevant to the query.  We closed this channel at the outset of the project by requiring that the provided evidence be a concatenation of verbatim passages from the corpus.  This requirement is documented to the meta-agent and enforced by the harness, which discards violating passages visibly, so that the meta-agent can appropriately revise future agents.

\paragraph{2. Unpriced resources become optimizer targets.}
AstaBench rewards a cost/accuracy Pareto curve.  However in addition to agent cost, there are additional resources which affect the practicality of the overall experimental campaign:  the cost of the LLM judge and wall-clock time. Both are ``free'' from the perspective of the evolutionary meta-agent and we observed evolution rationally exploiting these to optimize accuracy within the cost constraint, thus requiring a response on our side.

With respect to judge cost, an early \pfb{} run discovered that it could improve its F1 score by generating more evidence per paper, which increased the cost of the judge.  
Mean evidence grew to 1,257 characters per paper and p90 reached $\approx$4,700 characters.  We responded by imposing a 2,500-character limit, which we documented to evolution and enforced with a visible truncation. In the next comparable run, mean
evidence fell to 840 characters per paper and p90 to
$\approx$2,400---the evolved
agent engineered its \emph{own} truncation at 2,400 characters,
leaving a margin beneath the cap.

Based on our experience with DS-1000, we did not expect wall-clock time to be a binding constraint, so we documented the same 30-minute per-example limit for \pfb{} and we initially deviated from our ``document but don't editorialize'' policy by noting that we expected the limit to be non-binding.  This proved to be a mistake:  the evolutionary meta-agent rationally exploited the free wall-clock time by making repeated time-consuming calls to the Asta API and one agent lost 21 points to timeouts.  We responded by deleting the editorializing and simply documenting the limit.  

More subtly, \robophd{}
allows each evolution session 60 minutes, a limit we left out of the
prompt because sessions in earlier runs averaged
8--25 minutes. The first Opus-5 run changed that with no change to the
harness: its first ten sessions averaged 38 minutes and one hit the
60-minute limit, crashing the run. Adding one line to the evolution
prompt mid-run---``this session is capped at 60 minutes of wall
clock''---dropped the remaining sessions of the same run to a
10-minute average.

\paragraph{3. Appropriately pricing cost overage.}
As noted above, the evolutionary meta-agent is specifically instructed that its goal is to maximize the score on held-out queries and we give evolution a detailed chart showing how cost trades off against accuracy above the run's threshold, $\tau$, to yield a score (Table~\ref{tab:penalty}).   The term $\lambda$ in Eq.~\ref{eq:objective} sets the terms of that tradeoff and is thus a key parameter.  Initial experiments with a fixed dollar $\lambda$ (\$0.001--\$0.02, set per run) showed that the evolutionary meta-agent could exceed the cost threshold $\tau$ in order to improve its score when a lenient $\lambda$ made that arithmetic favorable. For instance, in one early run at $\lambda/\tau = 0.33$ the evolved agent exceeded $\tau$ by 14\%.
On the other hand, we worried that setting $\lambda$ too low would
lead to overly conservative behavior since exceeding the threshold by a
small margin would yield a catastrophic drop in score. We responded by
settling on $\lambda=0.1\tau$ and observe that across both campaigns, none of the twenty-three runs described in this paper with
$\lambda \le 0.1\tau$ shipped an agent which exceeded the cap (see Table~\ref{tab:slope}). Utilization also rises
almost monotonically with the strength of the evolution model, from
2.5--51\% of the cap under Opus-4.7 to 78--99\% under Fable-5
(Table~\ref{tab:slope}). The Fable models generally treat the cap as a
resource to be used, which is the intent. The top-scoring \dsk{} run with 43\% utilization is the exception: its winning agent
appeared at iteration~6 and finished first in 15 of its 17 training
rounds, so evolution saturated the training signal early
and had no reason to spend the remaining headroom.

\paragraph{4. The harness-improvement loop.}
At the end of an evolutionary iteration (after creating a new agent), evolution writes a ``reflection'' file which includes \emph{change
requests to the operator}.  For instance, meta-agent-requested changes to \pfb{} included providing diagnostics in JSON format, more complete API documentation, and hardening of the handling of calls to the Asta MCP API. 
Thus we can see a division of labor in which evolution improves the \emph{agent} while the operator improves the
\emph{evolutionary environment}, guided by evolution's requests. As future work, \robophd{} has an under-tested meta-evolution capability which may be able to automate this process.

\paragraph{5. Enabling evolutionary meta-agent testing and experimentation.}

One
of the most requested features in the reflections was the ability of the evolutionary
meta-agent to directly test calls to the Asta MCP API. Early runs could
only infer tool behavior from evaluation logs, and the reflections
tracked the gap explicitly: one session was rather emphatic---\emph{``Still no way to run against the live MCP server.
Sixth session, sixth request. Every agent shipped in this project has
never touched the real tools''}---and a session in a submitted run
declined a design change for want of \emph{``a 10-minute live probe,''} noting
that \emph{``a future instance with tool access should''} run it.
We responded in later runs with a well-documented session-side probe
granting sessions read-only access to the same tools available to
their agents (documentation reproduced in Appendix~\ref{app:inputs}).
Evolution used it heavily: 14 of 19 reflections in the
board leader's run reference the probe, crediting it with
\emph{``answer[ing] questions the docs left open, cheaply''}---one probe call
on an LLM-expanded title \emph{``returned exactly the gold id the seed
missed---validating the whole specific-query design in one call.''}

\section{Discussion}
\label{sec:discussion}

\paragraph{Evolutionary vs.\ inference costs.}
This work focuses on the cost of the solver agent, not the cost of the
evolutionary run that produced it (the meta-agent's own usage, training
evaluations, and, on \pfb{}, an LLM judge; Appendix~\ref{app:runs}
breaks these down). We do so for three reasons. First, solver cost is
the cost that scales: evolution is paid once per agent, while solver
cost is paid on every query the agent then serves. Second, judge cost is
a cost of measurement, which no deployment incurs. Third, solver cost is
the basis on which AstaBench ranks every entry, so it is the only basis
on which our agents can be compared with the rest of the board.

\paragraph{Pricing of other inference resources.}
This work took place within AstaBench, which prices only the solver
agent's LLM calls. Other inference-time resources, such as wall-clock
time, are unpriced
there and so are unpriced here; time was only limited by a
hard per-example cap (Section~\ref{sec:manage}, item~2). We note,
however, that Eq.~\ref{eq:objective} extends naturally to any metered
inference resource: mean wall-clock time, for instance, could be free
below its own threshold and penalized above it by a second term of the
same form. We leave this extension to future work.

\section{Related work and conclusion}
\label{sec:related}

\paragraph{Two lineages of agent optimization.}
  LLM-reflective optimizers split by what they let the optimizer change.
  A \emph{refinement} lineage holds the program fixed and optimizes one
  component around a strong incumbent: the prompt (GEPA
  \citep{gepa2025}) or, in a 2026 wave, the harness---the code that
  stores, retrieves, and presents context to a fixed model: Meta-Harness
  \citep{lee2026metaharness}, HARBOR \citep{sengupta2026harbor} (Bayesian
  optimization over harness flags), Self-Harness
  \citep{zhang2026selfharness}, and HarnessForge
  \citep{chen2026harnessforge}. \robophd{}
  \citep{robophd2026engine}, on the other hand, derives from a separate
  \emph{open-ended} lineage which begins with seeds that contain
  little more than an interface and which lets the whole artifact
evolve: AlphaEvolve \citep{alphaevolve2025,openevolve2025}, ADAS
  \citep{hu2025adas}, GEPA's optimize-anything mode \citep{gepa_oa2026},
  and Autoresearch \citep{autoresearch2026}. Meta-Harness, developed
  contemporaneously with \citet{robophd2026engine}, is the nearest
  neighbor by mechanism---a coding-agent proposer with filesystem access
  to prior candidates' code, scores, and traces---but not by lineage: its
  seeds are incumbents (ACE, Terminus-KIRA), its model is fixed, and its
  discovered harnesses build context for a fixed call pattern (one call
  per math problem; $\sim$80 lines atop Terminus-KIRA), staying close to
  the seed by what the authors call emergent strategy rather than rule.
  \citet{robophd2026engine}
  documented an ability for \robophd{} to evolve complex multi-call architectures
  (a 22-line ARC-AGI seed grew to 1{,}013 lines), but the model behind
  every call was predetermined by the framework; here the model joins the
  evolved artifact and cost joins the scoring function. The two are
  inseparable: without cost pressure, evolution would migrate to top-tier
  models; without a menu, evolution would find it more difficult to
  construct agents which fully exploit its available budget, as the
  ablation of Section~\ref{sec:pfb} shows.

  \paragraph{Which generalization?}
  The two lineages
  target different generalization goals. Refinement systems, having
  stayed near a generic incumbent, can claim \emph{artifact} transfer:
  Meta-Harness ports one harness to unseen datasets and different base models.
  Open-ended evolution specializes the artifact by design. 
  The goal of \robophd{} and other members of its lineage is to
  optimize a single artifact for a single task. Together with the four
  tasks of \citet{robophd2026engine}, the two reported here bring the
  evidence that the \robophd{} evolutionary meta-agent achieves this
  goal to six diverse tasks, each evolved from a modest seed with short
  background and objective markdown documents describing the task.  We make no claim that this work's agents
  transfer beyond their task, cost target, LLM menu, and tool
  suite. Our generalization claim is
  that the \robophd{} algorithm can evolve effective agents for a wide
  range of tasks, given modest task-specific information.

\paragraph{Cost-aware systems: where model choice lives.}
Prior systems place the which-model decision at one of three loci.
\emph{Outside the program}: FrugalGPT cascades and RouteLLM routers
\citep{chen2023frugalgpt,ong2024routellm} add dispatch logic around a
fixed scaffold---tuning spend along the curve the fixed program
defines, unable to restructure the program to be optimal \emph{at} a
requested price. \emph{As structured search over assignment}:
LLMSelector \citep{chen2025llmselector} allocates models to the modules
of a given pipeline by staged greedy search under an explicit
monotonicity assumption; Heterogeneous Swarms \citep{feng2025swarms}
searches topology and assignment jointly but over strictly feed-forward
DAGs; further instances restrict the domain or fix the topology family
\citep{syftr2025,bamas2025,si2025ccpo}. In every case a fixed
representation bounds what assignment can mean. For instance, a feed-forward DAG cannot express the
escalation, cross-provider consensus, and metadata-keyed routing our
board entries in fact use (Appendix~\ref{app:arch}). \emph{Inside the
evolved artifact}: to our knowledge \robophd{} is the first to combine a priced
multi-provider menu, free-form program evolution, and cost-penalized
selection.  This combination makes architecture and model
assignment a single design act rather than an allocation over a frozen
structure. On the objective side, the soft resource penalty descends from NAS
\citep{tan2019mnasnet}; GEPA's cost handling \citep{gepa_oa2026}---a
per-example budget with a fixed haircut---is the closest in-objective
prior, and our batch-average free zone and graded settable slope are
what turn the cost term into an operating-point dial
(Section~\ref{sec:system}).

\paragraph{Conclusion.}

We sequenced our work on this project by first implementing \dsk{} and
then taking on \pfb{}. Given the baseline of the \robophd{} engine
from \citet{robophd2026engine} and the model menu and cost work we
implemented for \dsk{}, the incremental work on \pfb{} was largely
focused on providing the right tools and documentation
(Section~\ref{sec:manage}, items~4--5). As noted above, this work led
us to take the top accuracy position on the board (at lower price)
from the manually constructed Asta Paper Finder system, and later to
dominate its cheaper configuration as well---a system for which
fourteen people were credited. This leads us to speculate that the
optimal user for \robophd{} might be a domain expert with modest
coding skills who could provide expert guidance to evolution through
high-quality background documents. This easy on-ramp, combined with
Pareto-dominance across two orders of magnitude on the cost axis,
suggests \robophd{} can serve as a broadly applicable platform.

\section*{Responsible-use statement}
The
supervision problems documented in Section~\ref{sec:manage} are the societal
risks of the approach in miniature: an evolutionary optimizer will fabricate
persuasive-but-ungrounded evidence if a judge rewards it (item~1), and
consume any unpriced shared resource (item~2). We report the incidents
alongside the guardrails that answered them.  For instance, the evidence grounding requirement prevented fabrications from 
reaching the judge (item~1). Appropriate resource caps that were disclosed to
the evolutionary meta-agent handled item~2. All leaderboard submissions were disclosed to the
maintainers as machine-generated agents on the leaderboard's ``Standard'' tooling and scored by
the benchmark's official evaluation; no benchmark-side systems were probed
beyond their public interfaces. We believe the net contribution is
defensive: an operating manual for running increasingly capable meta-agents
against shared evaluation infrastructure without corrupting it.

\section*{Reproducibility statement}
Appendix~\ref{app:boards} reproduces both leaderboards from the official
results dataset with true agent-only costs, snapshot-dated;
Appendix~\ref{app:runs} lists every submitted run's configuration
($\tau$, $\lambda$, evolution model), test scores, agent and
evolution costs, and cap utilization.
Per AstaBench policy, the underlying data are publicly available---the
maintainers publish every entry's raw results in the official dataset
(\texttt{allenai/asta-bench-results})---so our snapshots are
independently reconstructable. Headline numbers
are officially scored by the benchmark maintainers.
The evolutionary run behind every submitted agent is public at
\url{https://github.com/andborth/RoboPhD/tree/main/example_runs};
Table~\ref{tab:submissions} links to each one.
The engine is described in \citet{robophd2026engine}; the task
packages built here for the two tasks---objective and background
documents, seed agents, and the priced menu---are condensed in
Appendix~\ref{app:tasks} and released in full at
\url{https://github.com/andborth/RoboPhD}.

\section*{LLM usage}
There are five categories of LLM usage in this paper.  We use LLMs in the solver agents, in the evolutionary algorithm, and (in the case of the \pfb{} task) we use an LLM as a judge.  In addition, in keeping with the spirit of a project titled ``RoboPhD'', we note LLM usage in both the coding of the new task packages described herein and in manuscript preparation.  We document each in turn. 

Evolution/reflection models: Claude Opus 4.7, Claude Opus 4.8, Claude
Opus 5, Claude Fable 5, Claude Fable 5.1, and Claude Fable 5.2 running under Claude Code (per run; Appendix~\ref{app:runs}). Solver models: the nine-handle
menu of Appendix~\ref{app:tasks}. Judges: \texttt{gpt-4o-2024-11-20} (the
benchmark's official basis) for all reported \pfb{} scores;
\texttt{gpt-5.6-luna} (calibrated, Cohen's $\kappa=0.755$) for
training-time evaluation only, chosen for its more favorable cost
profile. Modifications to the \robophd{} engine and the two task packages described in this paper 
were almost entirely authored by Claude Code using Opus and Fable models under the
authors' direction.  A key exception was the background and objective documents described in Section~\ref{sec:system}, which incorporated significant human edits. Manuscript preparation balanced hand edits with closely supervised
Claude Opus and Claude Fable running under Claude Code.  All claims were verified against recorded run artifacts.

\begin{ack}
We thank Jonathan Bragg and the other AstaBench maintainers at Ai2 for
maintaining the leaderboards and reviewing our submissions. We also
thank Steve Ash and Anthony Galczak for their contributions to the
earlier \robophd{} work \citep{robophd2026engine} on which this paper
builds.
\end{ack}

\bibliographystyle{plainnat}
\bibliography{references}

\newpage
\appendix
\input{tables_boards}

\input{appendix}

\end{document}

%% file: tables_boards.tex
\newcommand{\dsboardrows}{%
\rowcolor{gray!12} \trophy{} & RoboPhD (Fable-5.1) & 0.873 & 0.054 & 0.127 & Claude-Opus-4-8, GPT-5.4, Gemini-3.1-Pro, +2 & Standard \\
\rowcolor{gray!12}  & RoboPhD (Opus-4.7) & 0.862 & 0.127 & 0.16 & Claude-Opus-4-7, Gemini-3.1-Pro, GPT-5.4, +2 & Standard \\
\rowcolor{gray!12}  & RoboPhD (Fable-5) & 0.853 & 0.052 & 0.06 & GPT-5.4, Claude-Sonnet-4-6, GPT-5.5, +1 & Standard \\
\rowcolor{gray!12} \trophy{} & RoboPhD (Opus-4.8) & 0.853 & 0.037 & 0.08 & GPT-5.4, Gemini-3.1-Pro, Claude-Sonnet-4-6 & Standard \\
 & ReAct & 0.849 & 0.247 & --- & Gemini-3.1-Pro & Standard \\
 & ReAct & 0.847 & 0.050 & --- & GPT-5.5 & Standard \\
 & ReAct & 0.838 & 0.044 & --- & GPT-5.4 & Standard \\
 & ReAct & 0.837 & 0.045 & --- & Claude-Opus-4-6 & Standard \\
\trophy{} & Button & 0.836 & 0.028 & --- & Claude-Opus-4-6 & Custom \\
\rowcolor{gray!12} \trophy{} & RoboPhD (Opus-4.8) & 0.830 & 0.017 & 0.05 & GPT-5.4, Claude-Sonnet-4-6, GPT-5.4-Mini & Standard \\
\rowcolor{gray!12} \trophy{} & RoboPhD (Opus-4.7) & 0.809 & 0.012 & 0.08 & Claude-Sonnet-4-6 & Standard \\
 & ReAct & 0.786 & 0.058 & --- & Claude-Opus-4-7 & Standard \\
 & EvoScientist-Code & 0.784 & 0.033 & --- & GPT-5 & Custom \\
 & ReAct & 0.780 & 0.021 & --- & GPT-5 & Standard \\
 & Smolagents Coder & 0.757 & 0.018 & --- & GPT-5 & Custom \\
 & ReAct & 0.756 & 0.044 & --- & Claude-Sonnet-4 & Standard \\
\rowcolor{gray!12} \trophy{} & RoboPhD (Opus-4.8) & 0.751 & 0.007 & 0.05 & GPT-5.4-Mini, GPT-5.4 & Standard \\
 & ReAct & 0.749 & 0.010 & --- & o3 & Standard \\
 & Asta v0 & 0.748 & 0.011 & --- & Claude-Sonnet-4, Gemini-2.0-Flash, o3, +2 & Full \\
 & Smolagents Coder & 0.747 & 0.114 & --- & Claude-Sonnet-4 & Custom \\
 & ReAct & 0.741 & 0.040 & --- & Claude-Sonnet-4-6 & Standard \\
\rowcolor{gray!12} \trophy{} & RoboPhD (Opus-4.8) & 0.737 & 0.002 & 0.003 & GPT-5.4, GPT-5.4-Mini & Standard \\
\rowcolor{gray!12}  & RoboPhD (Fable-5) & 0.732 & 0.004 & 0.003 & Gemini-3.1-Flash-Lite, GPT-5.4-Mini, Claude-Haiku-4.5, +1 & Standard \\
 & ReAct & 0.710 & 0.003 & --- & GPT-5-Mini & Standard \\
 & ReAct & 0.670 & 0.007 & --- & GPT-4.1 & Standard \\
 & Smolagents Coder & 0.652 & 0.016 & --- & GPT-5-Mini & Custom \\
 & ReAct & 0.554 & 0.019 & --- & Gemini-2.5-Flash & Standard \\
 & ReAct & 0.541 & 0.005 & --- & Claude-3.5-Haiku & Standard \\
 & Smolagents Coder & 0.480 & 0.073 & --- & GPT-4.1 & Custom \\
 & ReAct & 0.437 & 0.010 & --- & GPT-4o & Standard \\
 & Smolagents Coder & 0.289 & 0.044 & --- & Gemini-2.5-Flash & Custom \\
 & Smolagents Coder & 0.168 & 0.137 & --- & GPT-4o & Custom \\
 & Smolagents Coder & 0.099 & 0.024 & --- & Claude-3.5-Haiku & Custom \\
 & ReAct & 0.097 & 0.110 & --- & Llama-4-Scout & Standard \\
 & Smolagents Coder & 0.027 & 0.004 & --- & Llama-4-Scout & Custom \\}
\newcommand{\pfboardrows}{%
\rowcolor{gray!12} \trophy{} & RoboPhD (Fable-5.2) & 0.501 & 0.232 & 0.279 & GPT-5.4-Mini, GPT-5.4, Claude-Opus-4-8, +3 & Standard \\
\rowcolor{gray!12}  & RoboPhD (Fable-5) & 0.440 & 0.279 & 0.355 & GPT-5.4, GPT-5.4-Mini & Standard \\
 & Asta Paper Finder & 0.433 & 0.355 & --- & GPT-5-Mini, GPT-4o, Gemini-3-Flash & Custom \\
\rowcolor{gray!12}  & RoboPhD (Opus-5) & 0.432 & 0.251 & 0.355 & GPT-5.4-Mini, Claude-Sonnet-4-6, GPT-5.4 & Standard \\
\rowcolor{gray!12} \trophy{} & RoboPhD (Fable-5.1) & 0.399 & 0.058 & 0.063 & GPT-5.4-Mini, GPT-5.4, Claude-Sonnet-4-6 & Standard \\
 & Asta Paper Finder & 0.397 & 0.063 & --- & Gemini-2.0-Flash, GPT-4o & Custom \\
\rowcolor{gray!12} \trophy{} & RoboPhD (Fable-5) & 0.390 & 0.058 & 0.063 & GPT-5.4-Mini, GPT-5.4 & Standard \\
\rowcolor{gray!12} \trophy{} & RoboPhD (Opus-5) & 0.376 & 0.052 & 0.063 & GPT-5.4-Mini, Claude-Haiku-4.5, GPT-5.4, +1 & Standard \\
 & Asta v0 & 0.376 & 0.063 & --- & Claude-Sonnet-4, Gemini-2.0-Flash, o3, +2 & Full \\
\rowcolor{gray!12}  & RoboPhD (Fable-5) & 0.375 & 0.053 & 0.06 & GPT-5.4-Mini, GPT-5.4 & Standard \\
 & ReAct & 0.374 & 3.381 & --- & Claude-Opus-4-7 & Standard \\
 & ReAct & 0.372 & 1.489 & --- & Claude-Opus-4-6 & Standard \\
 & ReAct & 0.360 & 7.158 & --- & GPT-5.5 & Standard \\
 & ReAct & 0.356 & 2.395 & --- & Claude-Sonnet-4-6 & Standard \\
 & ReAct & 0.301 & 2.023 & --- & GPT-5.4 & Standard \\
 & ReAct & 0.264 & 2.257 & --- & Gemini-3.1-Pro & Standard \\
 & ReAct & 0.264 & 0.428 & --- & GPT-5 & Standard \\
 & Smolagents Coder & 0.221 & 0.975 & --- & Claude-Sonnet-4 & Custom \\
\rowcolor{gray!12} \trophy{} & RoboPhD (Opus-4.8) & 0.220 & 0.006 & 0.033 & GPT-5.4-Mini & Standard \\
 & ReAct & 0.220 & 0.060 & --- & GPT-5-Mini & Standard \\
 & ReAct & 0.203 & 0.505 & --- & Claude-Sonnet-4 & Standard \\
 & Smolagents Coder & 0.200 & 0.120 & --- & GPT-5 & Custom \\
 & ReAct & 0.193 & 0.518 & --- & o3 & Standard \\
 & Smolagents Coder & 0.172 & 0.033 & --- & GPT-5-Mini & Custom \\
 & Smolagents Coder & 0.165 & 0.079 & --- & GPT-4.1 & Custom \\
 & ReAct & 0.165 & 0.867 & --- & GPT-4.1 & Standard \\
 & Smolagents Coder & 0.147 & 0.044 & --- & Gemini-2.5-Flash & Custom \\
 & ReAct & 0.129 & 0.267 & --- & GPT-4o & Standard \\
 & Smolagents Coder & 0.125 & 0.098 & --- & GPT-4o & Custom \\
 & ReAct & 0.107 & 0.060 & --- & Claude-3.5-Haiku & Standard \\
 & Smolagents Coder & 0.070 & 0.013 & --- & Llama-4-Scout & Custom \\
 & ReAct & 0.065 & 1.196 & --- & Gemini-2.5-Flash & Standard \\
 & ReAct & 0.054 & 2.815 & --- & Llama-4-Scout & Standard \\
 & Smolagents Coder & 0.046 & 0.070 & --- & Claude-3.5-Haiku & Custom \\}

%% file: appendix.tex
\section{Leaderboard snapshots}
\label{app:boards}

Both boards are from the official AstaBench results dataset
(\texttt{allenai/asta-bench-results}, test split), snapshot
\snapdate{}; costs are the dataset's agent-only per-problem
means, reported to the nearest tenth of a cent (scores to three
decimals), the paper's precision convention. Raw full-precision
submission files are archived with the paper materials. Pareto dominance
is adjudicated on full-precision point estimates, with score ties
resolved to the cheaper entry---the leaderboard's official semantics.
Two \pfb{} claims turn on digits below the precision shown: the \$0.006
entry over ReAct/GPT-5-Mini (0.22046 vs.\ 0.22039, at a tenth of the
cost) and the \$0.052 entry over Asta v0 (0.3762 vs.\ 0.3757, 16\%
cheaper). The live boards
can be found at \url{https://allenai-asta-bench-leaderboard.hf.space/}:
Literature Understanding $\to$ \pfb{} and Code \& Execution $\to$ \dsk{}.

\begin{table}[h]
\centering
\scriptsize
\setlength{\tabcolsep}{4pt}
\begin{tabular}{@{}clrrrll@{}}
\toprule
 & Agent & Score & \$/problem & $\tau$ (\$) & Model & Tools \\
\midrule
\dsboardrows
\bottomrule
\end{tabular}
\caption{\dsk{} board, snapshot \snapdate{} (shaded = ours). The
parenthetical on each \robophd{} row names the \emph{evolution} model
that built that agent (per-run details in Table~\ref{tab:runs}); the
evolution model is never called at inference. $\tau$ is the
operator-chosen per-problem cost target the agent was evolved to
(Section~\ref{sec:system}); entries in the other classes have no such
parameter. \trophy{} marks the entries on the board's cost--quality
Pareto frontier (the staircase of Figure~\ref{fig:money}); on this board,
exactly one frontier entry (Button) is not a \robophd{} agent.
Model lists are the models each entry actually called, by share of tokens
(see Table~\ref{tab:pfboard}).}
\label{tab:dsboard}
\end{table}

\begin{table}[h]
\centering
\scriptsize
\setlength{\tabcolsep}{4pt}
\begin{tabular}{@{}clrrrll@{}}
\toprule
 & Agent & Score & \$/problem & $\tau$ (\$) & Model & Tools \\
\midrule
\pfboardrows
\bottomrule
\end{tabular}
\caption{\pfb{} board, snapshot \snapdate{} (shaded = ours; \robophd{}
parentheticals, the $\tau$ column, and the \trophy{} frontier marker as
in Table~\ref{tab:dsboard}). Every frontier entry on this board is a
\robophd{} agent.
\textbf{Model} lists the models an entry actually called, recovered from
the per-sample \texttt{model\_usages} in its submission file and ordered by
share of total tokens.}
\label{tab:pfboard}
\end{table}

\section{Run registry}
\label{app:runs}

Table~\ref{tab:runs} lists every submitted run; Table~\ref{tab:slope} lists the
twenty-three runs at $\lambda \le 0.1\tau$ behind Section~\ref{sec:manage},
item~3, fourteen of which are unsubmitted. ``Util.''\ is achieved mean agent cost over threshold $\tau$---the
precision metric for cost targeting. Evolution cost is the meta-loop LLM
spend (per-iteration metering); run total additionally includes
training evaluation and, for \pfb{}, training judge spend.

\begin{table}[h]
\centering
\scriptsize
\setlength{\tabcolsep}{3.5pt}
\begin{tabular}{clllrrrrr}
\toprule
 & Task & Evol.\ model & $\tau$ / $\lambda$ (\$) & Test & \$/prob & Util. & Evol.\ \$ & Agent lines \\
\midrule
\trophy{} & \dsk{} & Fable-5.1 & 0.127 / 0.0127 & 0.873 & 0.054 & 43\% & 167.10 & 1{,}226 \\
 & \dsk{} & Opus-4.7 & 0.16 / ---$^{\dagger}$ & 0.862 & 0.127 & 80\% & 45.44 & 1{,}214 \\
 & \dsk{} & Fable-5 & 0.06 / 0.01 & 0.853 & 0.052 & 87\% & 64.09 & 1{,}614 \\
\trophy{} & \dsk{} & Opus-4.8 & 0.08 / 0.01 & 0.853 & 0.037 & 46\% & 23.13 & 500 \\
\trophy{} & \dsk{} & Opus-4.8 & 0.05 / 0.01 & 0.830 & 0.017 & 34\% & 31.67 & 525 \\
\trophy{} & \dsk{} & Opus-4.7 & 0.08 / $\approx$50$^{\ddagger}$ & 0.809 & 0.012 & 15\% & 24.17 & 561 \\
\trophy{} & \dsk{} & Opus-4.8 & 0.05 / 0.01 & 0.751$^{\P}$ & 0.007 & 14\% & 32.52 & 354 \\
\trophy{} & \dsk{} & Opus-4.8 & 0.003 / 0.0003 & 0.737 & 0.002 & 56\% & 33.07 & 302 \\
 & \dsk{} & Fable-5 & 0.003 / 0.001 & 0.732 & 0.004 & 143\%$^{\S}$ & 153.97 & 1{,}419 \\
\midrule
\trophy{} & \pfb{} & Fable-5.2 & 0.279 / 0.0279 & 0.501 & 0.232 & 83\% & 122.23 & 1{,}811 \\
 & \pfb{} & Fable-5 & 0.355 / 0.0355 & 0.440 & 0.279 & 78\% & 142.71 & 2{,}781 \\
 & \pfb{} & Opus-5 & 0.355 / 0.0355 & 0.432 & 0.251 & 71\% & 73.00 & 1{,}884 \\
\trophy{} & \pfb{} & Fable-5.1 & 0.063 / 0.0063 & 0.399 & 0.058 & 92\% & 203.74 & 2{,}088 \\
\trophy{} & \pfb{} & Fable-5 & 0.063 / 0.0063 & 0.390 & 0.058 & 92\% & 138.18 & 2{,}068 \\
\trophy{} & \pfb{} & Opus-5 & 0.063 / 0.0063 & 0.376 & 0.052 & 83\% & 72.11 & 2{,}349 \\
 & \pfb{} & Fable-5 & 0.06 / 0.02 & 0.375 & 0.053 & 89\% & 204.10 & 2{,}097 \\
\trophy{} & \pfb{} & Opus-4.8 & 0.033 / 0.003 & 0.220 & 0.006 & 18\% & 64.89 & 1{,}006 \\
\bottomrule
\end{tabular}
\caption{Run registry: the seventeen submitted runs, in the board order of Table~\ref{tab:submissions}. \trophy{}: on the board's Pareto frontier, as in Table~\ref{tab:submissions}. $^{\dagger}$Included for completeness as a submitted run. It
predates Eq.~\ref{eq:objective}: cost was penalized per example above
$\tau$, as in prior work \citep{gepa_oa2026}.
$^{\ddagger}$Used Eq.~\ref{eq:objective}, but with a very large
$\lambda$: about \$50. Hence cost acted as a tiebreaker between agents
tied on accuracy.
$^{\P}$The registry's one Deep Focus run---a rerun of the \$0.017
run's configuration, matched
on every setting ($\tau$, $\lambda$, model, evaluation budget) except
enabling the engine's Deep Focus refinement
(\citealp{robophd2026engine}, \S3.2), which every other listed run
disables. $^{\S}$Utilization on the official billing basis; high because an
error in the Gemini billing reported to the evolutionary model led it
to overspend.}
\label{tab:runs}
\end{table}

\input{tables_slope}

\begin{table}[h]
\centering
\small
\setlength{\tabcolsep}{4.5pt}
\begin{tabular}{@{}lrrrrrrrr@{}}
\toprule
 & & \multicolumn{3}{c}{Score by query type} & & & & \\
\cmidrule(lr){3-5}
Menu & Lines & Semantic & Specific & Metadata & All & $\Delta$ (s.e.) & \$/query & Util. \\
 & & (194) & (38) & (35) & (267) & & & \\
\midrule
Full (nine models) & 1{,}811 & 0.459 & 0.820 & 0.461 & 0.510 & --- & 0.232 & 83\% \\
GPT-5.4-mini only & 1{,}521 & 0.446 & 0.838 & 0.378 & 0.493 & $-0.018$ (0.011) & 0.203 & 73\% \\
GPT-5.4 only & 1{,}664 & 0.419 & 0.811 & 0.297 & 0.459 & $-0.051$ (0.012) & 0.205 & 73\% \\
\bottomrule
\end{tabular}
\caption{Model-menu ablation on \pfb{} at $\tau = \$0.279$
(Section~\ref{sec:pfb}): three evolutionary runs that differ only in the
menu offered. The full-menu run is the board leader. Scores
are on the 267 held-out queries and luna-judged
(see Appendix~\ref{app:tasks}), so the leader reads 0.510 here against 0.501
on the official basis.}
\label{tab:menuablation}
\end{table}

\paragraph{Cost anatomy.}
Campaign totals decompose into three categories that are metered
separately in our records and have different practical characters:
\emph{evolution} spend (the meta-loop model's own calls),
\emph{training-evaluation} spend (running candidate agents on the
training pool, on the same priced menu as deployment), and
\emph{training-judge} spend (\pfb{} only). One worked example, the
\pfb{} \$0.355 Fable-5 arm (the \$0.279 board entry): training cost \$275.38
all-in---\$142.71 evolution, \$124.51 training evaluation, and \$8.16
training judge (luna). Measuring the result then cost \$144.58 for the
internal test evaluation on the official \texttt{gpt-4o} judge basis
(\$74.22 agent, \$70.36 judge) and \$277.99 for the official submission
evaluation (\$74.38 agent, \$203.61 judge, uncapped)---scoring the
submission cost more than the training run itself. The distinction matters
because the categories are procured differently in practice: evaluation
and judge spend are metered API calls---out-of-pocket dollars---while
the evolution model commonly runs under a subscription seat (an Anthropic subscription in our case with weekly usage limits), where its spend consumes
plan quota rather than incurring marginal dollars. On that accounting,
the worked example's \$275 training total splits into \$133 out of
pocket plus \$143 of subscription quota (measurement spend is entirely
out of pocket), and the \dsk{} runs' out-of-pocket
component is their training evaluation alone, \$1--\$41 across the
submitted runs. Evolution spend at list price varied with both the task and the
evolution model (Table~\ref{tab:runs}, ``Evol.~\$''): on \dsk{},
\$23--\$45 per run for the six Opus-evolved submissions and
\$64--\$167 for the three Fable-evolved ones; on \pfb{},
\$65--\$73 for the three Opus-evolved submissions and
\$122--\$204 for the five Fable-evolved ones.

\section{Task specifications and the model menu}
\label{app:tasks}

Both tasks are AstaBench \citep{bragg2026astabench} tasks on their
\emph{Standard} tooling tier. What follows is condensed from the actual
inputs given to evolution: a 5-line objective document and an API-reference
background document per task. By house rule the background documents
contain no strategy hints or recommended settings; they are essentially
transcribed task and API documentation, so the truly authored human input
per task is the $\sim$200-word objective and a minimal seed agent.
Appendix~\ref{app:inputs} reproduces the instantiated documents from
two submitted runs---objectives in full and backgrounds in verbatim
excerpt---plus the \dsk{} seed verbatim and the \pfb{} seed as
pseudocode.

\subsection{The model menu}
\label{app:menu}

Agents may call LLMs only through nine pre-resolved handles
(Table~\ref{tab:menu}); prices are the rates the benchmark's scoring
bills. Each handle accepts a per-call
\texttt{reasoning\_effort} override and a \texttt{max\_tokens} cap. The
handle layer rides on Inspect \citep{inspectai2024}, the UK AI Security
Institute's evaluation framework on which AstaBench itself is built: each
handle resolves to an \texttt{inspect\_ai} model object, and the per-call
overrides are Inspect \texttt{GenerateConfig} fields---so evolved agents
are ordinary Inspect solvers, submissible to the leaderboard unmodified.

\begin{table}[h]
\centering
\small
\begin{tabular}{lrrl}
\toprule
Handle & Input \$/Mtok & Output \$/Mtok & Default reasoning \\
\midrule
GPT-5.4-Mini & 0.75 & 4.50 & none \\
GPT-5.4 & 2.50 & 15.00 & none \\
GPT-5.5 & 5.00 & 30.00 & model-managed \\
Claude-Haiku-4.5 & 1.00 & 5.00 & none \\
Claude-Sonnet-4.6 & 3.00 & 15.00 & none \\
Claude-Opus-4.8 & 5.00 & 25.00 & model-managed \\
Gemini-3.1-Flash-Lite & 0.45 & 2.70 & low \\
Gemini-3.5-Flash & 1.50 & 9.00 & low \\
Gemini-3.1-Pro-Preview & 2.00 & 12.00 & low \\
\bottomrule
\end{tabular}
\caption{The priced menu offered to the evolutionary meta-agent. Roughly $10\times$ span in output price; three
providers; cheap/standard/strong tiers per provider.}
\label{tab:menu}
\end{table}

\subsection{\dsk{}}
\textbf{Objective (verbatim opening and closing).} \emph{``Evolve a DS-1000
agent that, given a Python data-science problem prompt, produces a
\texttt{<code>...</code>} block whose contents make the hidden test
program's \texttt{result} variable match the reference under all hidden
test inputs. [\dots] Each iteration draws a different sample of problems,
and the final agent is evaluated on a held-out test set it has never seen.
So, to rephrase your goal, your objective is to build an agent that
generalizes to unseen problems---the visible batch is a training signal,
not the target.''}

\textbf{Data and scoring.} 100-problem visible pool; 900-problem held-out
test. Binary per-problem scoring by hidden execution tests; a subset of
problems additionally enforce style/idiom constraints on the submitted code
(e.g.\ forbidding explicit loops, or requiring a named library function),
surfaced to the agent only through assertion tracebacks. Batch score
follows Eq.~\ref{eq:objective}: correct answers minus cost-overage
error-equivalents, scaled to 0--100. Only
menu calls are metered; the Python sandbox is free. Per-example wall-clock
cap 1{,}800\,s (documented to evolution as 29 minutes, leaving a margin).

\subsection{\pfb{}}
\textbf{Objective (verbatim opening and closing).} \emph{``Evolve a
PaperFindingBench agent that, given a natural-language literature-search
query, returns a list of Semantic Scholar corpus\_ids maximizing adjusted
micro-F1 against the query's hidden gold, using only the Standard tools
(Asta MCP corpus + model\_registry LLM handles). [\dots] your objective is
to build an agent that generalizes to unseen queries---the visible batch is
a training signal, not the target.''}

\textbf{Data and scoring.} 66-query visible pool; 267-query held-out test.
The benchmark's own \texttt{query\_id} prefix assigns each query to one of
three classes, and the class selects the scorer:

\begin{itemize}
\item \texttt{specific} (38/267): the query names a single paper the user
  already has in mind (``Find me the `attention is all you need' paper'';
  ``Find me the original GPT 3 paper''). Gold is a single work, listed as one
  corpus ID in 34 of the 38 test queries and as two or three duplicate
  records of the same paper in the remainder. Scored \texttt{specific\_f1}:
  harmonic mean of precision and recall over the intersection of submitted
  and gold IDs. Extra papers cost precision, so the target behavior is to
  return only the identified work.
\item \texttt{metadata} (35/267): the query fixes a set by bibliographic
  constraint---author, year, venue, publication type---rather than by topic
  (``2012 papers by David Harel''; ``Journal papers co-authored by David
  Harel and Shahar Maoz''; ``1988 papers by the author of the Statecharts
  paper''). Gold is an enumerated ID set, taken to be complete, so the same
  exact-match \texttt{metadata\_f1} applies and both over- and
  under-retrieval cost score.
\item \texttt{semantic} (194/267): the query states a topical information
  need (``Are there any tools or studies that have focused on building a
  morphological analyzer specifically for handling multiple Arabic
  dialects?''). Gold here is acknowledged to be incomplete---a small
  \texttt{known\_to\_be\_good} seed set plus weighted natural-language
  \emph{relevance criteria}---so scoring is TREC-style pooling with an LLM
  assessor: each submitted paper not already in gold is graded by the judge,
  and \texttt{semantic\_f1} is the harmonic mean of a rank term
  (lower-bound-corrected nDCG over those grades) and a recall term whose
  numerator counts only submissions the judge grades perfectly relevant and
  whose denominator is a per-query shipped estimate of the total relevant
  population---so recall is never an intersection with gold. The judge
  sees only the agent-supplied \texttt{markdown\_evidence} for each returned
  paper---the property Section~\ref{sec:manage} (item~1) hardens against.
\end{itemize}

\noindent The headline metric is the micro-average of per-query scores across
all three classes (\texttt{adjusted\_f1\_micro\_avg}). Tool calls (Asta MCP
corpus API) are free; only menu LLM calls are metered. Per-example
wall-clock cap 1{,}800\,s (documented to evolution as 29 minutes, leaving a margin).

\textbf{Grounding check (reward-channel hardening, main-text item 1).}
Before any judge call, every evidence passage is checked for verbatim
derivability (after markdown normalization) from corpus text the agent
actually retrieved through MCP tools \emph{in the same evaluation};
ungrounded passages are discarded pre-judge, so fabricated evidence scores
as an empty submission. The check is implemented evaluator-side and is not
visible to, or modifiable by, the evolved agent.

\textbf{Judge basis and calibration.} Two judges appear in this work:
the benchmark's official \texttt{gpt-4o-2024-11-20} and the cheaper
\texttt{gpt-5.6-luna}, run with a no-prose prompt, which we use where
cost requires it. \emph{In training}, all submitted runs except the
first (\pfb{}'s \$0.053 entry) were trained against luna. \emph{At
test}, every reported \pfb{} score---and every submission---is judged
by \texttt{gpt-4o}, with two exceptions: the waypoint scores of
Table~\ref{tab:waypoints} and Figure~\ref{fig:waypoints} and the
model-menu ablation of Table~\ref{tab:menuablation} are judged by
luna throughout. No luna-judged number carries a board claim.
Luna was calibrated against \texttt{gpt-4o} on a paired sample (Cohen's
$\kappa = 0.755$) and costs roughly an order of magnitude less. Because the two judges agree
imperfectly, luna-trained evolution partly optimizes a proxy: selection
credit earned under luna does not always survive re-judging by
\texttt{gpt-4o}. This train/score mismatch is a systematic headwind on
our reported \pfb{} results---the submitted agents were tuned against a
judge they are not scored by---accepted as the price of affordable
training-scale judging.

\subsection{Example problems}
\label{app:examples}

All examples below are drawn from the \emph{visible} training pools; no
held-out test items are reproduced.

\paragraph{\dsk{} (visible-pool problem 34, Pandas).}
The agent receives the question, the given code context, and the
completion instruction:

\begin{quote}\small
\emph{``I have a script that generates a pandas data frame with a varying
number of value columns. [\dots] My goal is to get the grouped sum for
each of the value columns. In this specific case (with 2 value columns), I
can use} \texttt{df.groupby('group').agg(\{"group\_color": "first",
"val1": "sum", "val2": "sum"\})} \emph{but that does not work when the
data frame in question has more value columns (val3, val4 etc.). Is there
a way to dynamically take the sum of ``all the other columns'' or ``all
columns containing val in their names''?''}
\end{quote}

\begin{verbatim}
A:
<code>
import pandas as pd
df = pd.DataFrame({ 'group': ['A', 'A', 'A', 'B', 'B'],
  'group_color' : ['green', 'green', 'green', 'blue', 'blue'],
  'val1': [5, 2, 3, 4, 5], 'val2' : [4, 2, 8, 5, 7],
  'val3':[1,1,4,5,1] })
</code>
result = ... # put solution in this variable
BEGIN SOLUTION
\end{verbatim}

The agent must return only the completion, e.g.\ the reference solution:

\begin{verbatim}
def g(df):
    return df.groupby('group').agg(
        lambda x: x.head(1) if x.dtype=='object' else x.sum())
result = g(df.copy())
\end{verbatim}

Scoring executes the completed program against hidden test inputs (and,
for a subset of problems, style assertions on the submitted code itself);
the agent never sees the tests.

\paragraph{\pfb{} (visible-pool queries).}
Queries range from focused to colloquial. Four from the training pool:

\begin{quote}\small
\textbf{semantic\_100:} \emph{``What are some of the key advantages and
challenges of building multilingual evaluation datasets by translating
existing English datasets into other target languages? Or more
specifically, how do these translated datasets impact the quality and
reliability of cross-lingual evaluations, and what potential pitfalls
should be considered when using translations as a primary method for
creating non-English benchmarks?''}

\textbf{semantic\_108:} \emph{``Has anyone tried fine tuning with
RAG---or have seen work on this?''}

\textbf{metadata\_25:} \emph{``Papers citing the DistilBERT paper after
2022 with more than 50 citations''}

\textbf{specific\_11:} \emph{``the paper about the Objaverse dataset''}
\end{quote}

The agent returns ranked Semantic Scholar
\citep{kinney2023semanticscholar} corpus IDs, each with a
\texttt{markdown\_evidence} passage (grounding-checked,
Appendix~\ref{app:tasks}). For \texttt{specific} queries the gold is an
exact corpus-ID match (specific\_11 $\to$ one ID); for \texttt{metadata}
queries, intersection with an enumerated gold set (metadata\_25 $\to$
172 IDs). For semantic queries a
hidden weighted rubric drives the LLM judge; semantic\_100's gold
criteria, never shown to the agent, are: \emph{translation of English
datasets for multilingual evaluation} (weight 0.4), \emph{impact on
cross-lingual evaluation quality and reliability} (0.3), and
\emph{challenges and pitfalls of using translations} (0.3) --- the judge
assesses each returned paper against these using only the agent-supplied
evidence.

\section{Scores, costs, and agent reasoning indicating an absence of overfitting}
\label{app:waypoints}

As noted in Section~\ref{sec:overfit}, a key characteristic of overfitting is that we would see scores on held-out test data peaking and then declining over the course of an evolutionary run as the algorithm fit excessively to the available training data.  We present evidence to the contrary in Table~\ref{tab:waypoints}.  This table shows runs on each of our two tasks at two different price points.  The ``winner'' column in each case is the shipped agent of a run
displayed in Tables~\ref{tab:dsboard} and~\ref{tab:pfboard},
re-evaluated here internally so that all five waypoints share one
measurement basis; the board's official figures for the same agents
differ by evaluation draw and, for multi-model \dsk{} agents, by cost
meter (ours reads 70--89\% of the board's on those agents).  Along with each winner, we show the corresponding seed agent, the agent evolved in iteration 2 (the first evolved agent---which isolates the effect of Claude Code taking a single pass at the task with the benefit of one round of  seed agent results, but without the benefit of evolutionary selection), along with the Elo leaders as of iterations 6 and 11 (the agents which would have been produced had evolution stopped at those steps).  We see step over step increases in score in every case.  The other two metrics shown support the idea that this is no accident.  Lines of code and cost, similarly increase in every case, showing successively greater agent complexity and greater usage of the available headroom that our cost cap, $\tau$,
affords.

\begin{table}[h]
\centering
\small
\begin{tabular}{llrrrrr}
\toprule
Arm & & seed & iter.\ 2 & as-of-6 & as-of-11 & winner \\
\midrule
\pfb{} \$0.355 & score & 0.0476 & 0.2550 & 0.4214 & 0.4237 & 0.4703 \\
 & \$/query & 0.0006 & 0.0546 & 0.1663 & 0.1943 & 0.2780 \\
 & code lines & 72 & 560 & 1{,}156 & 1{,}419 & 1{,}861 \\
\pfb{} \$0.063 & score & 0.0476 & 0.3528 & 0.3953 & 0.4060 & 0.4137 \\
 & \$/query & 0.0006 & 0.0268 & 0.0408 & 0.0562 & 0.0583 \\
 & code lines & 72 & 501 & 973 & 1{,}410 & 1{,}489 \\
\dsk{} \$0.16 & score & 0.6914 & 0.7211 & 0.8229 & \emph{= winner} & 0.8629 \\
 & \$/prob. & 0.0005 & 0.0117 & 0.0369 & --- & 0.1008 \\
 & code lines & 20 & 261 & 455 & --- & 1{,}062 \\
\dsk{} \$0.08 & score & 0.6944 & 0.7633 & 0.8244 & 0.8322 & 0.8444 \\
 & \$/prob. & 0.0005 & 0.0016 & 0.0132 & 0.0236 & 0.0295 \\
 & code lines & 20 & 144 & 209 & 289 & 337 \\
\bottomrule
\end{tabular}
\caption{Held-out score, agent-only cost, and code size at each
waypoint. Every step increases on all three tracks (15/15). All
figures are internal evaluations on a single basis, so steps compare
within an arm; winners therefore differ from the board's official
figures for the same agents (see text). \pfb{} scores are on the
luna judge basis ($\approx$0.03 above the leaderboard's; Appendix~\ref{app:tasks}); \dsk{}
scoring is execution against hidden tests on the full 900-problem
held-out set. The
\dsk{} \$0.16 arm's as-of-11 leader is already its winner. Code lines
exclude docstrings, comments, and blank lines, in contrast to
Table~\ref{tab:runs}.}
\label{tab:waypoints}
\end{table}

Table~\ref{tab:mechanisms} tracks one run per task---the \pfb{}
\$0.355 arm and the \dsk{} \$0.08 arm---pairing each step with what the
evolution sessions implemented and, quoted from the producing
iteration's own records, why they expected it to transfer. The pattern
is uniform: each step attacks the dominant \emph{residual failure
class} with a mechanism generic to that class, and the class-level
structure is what the held-out gains inherit. The two untracked arms
follow the same failure-class discipline through different mechanisms
(citation-graph expansion and a title-guess channel on the \pfb{}
\$0.063 arm; sandbox-verified loop rewrites on the \dsk{} \$0.16 arm).

\begin{table}[h]
\centering
\scriptsize
\setlength{\tabcolsep}{4pt}
\begin{tabular}{p{0.105\textwidth}p{0.43\textwidth}p{0.40\textwidth}}
\toprule
Step & \pfb{} \$0.355 arm & \dsk{} \$0.08 arm \\
\midrule
seed$\to$iter.\ 2 \newline (the big jumps) &
Score-function-aware structure: route on the query-class label; submit
ranked lists up to 250 deep against the recall denominator
(``submitting a well-ordered list of up to 250 candidates is nearly
pure upside''); metadata queries moved onto the author/citation tools;
specific queries submit 1--2 papers, not eight (``with gold size
$\sim$1, submitting 8 papers caps F1 at 0.22 even on a hit'').
``Every failure is structural, not incidental.'' &
Execution grounding: run every candidate in the unmetered sandbox,
escalate models on failure---``we detect them for free
(\texttt{python\_session} is not metered) and can retry with the
traceback.'' Bought $+0.069$ held-out for $+$\$0.001/problem. \\
\addlinespace
iter.\ 2 $\to$ \newline as-of-6 &
Attack the next class: grade-2$\to$3 evidence repair and a
judge-mimicking second grading pass, once diagnostics showed recall
``the binding term on every single query.'' &
Attack the next class: ``code runs cleanly but the answer is subtly
wrong---exactly the class \texttt{verify\_escalate} cannot catch'';
a cross-family ensemble with sandbox output values compared and an
output-grounded judge---``generic to \dsk{}, not tuned to these 20.'' \\
\addlinespace
as-of-6$\to$ as-of-11 \newline (the dead zone) &
Lever saturation, tracked explicitly: ``six sessions of history say
semantic prompt tweaks are noise at this sample size [\ldots] the
deterministic metadata fixes carry the iteration''---the category
rotation of Section~\ref{sec:overfit}. &
``Candidate count is saturated; the remaining errors are not `no
candidate was right' errors''---judge-quality upgrades only. \\
\addlinespace
$\to$winner &
Channel widening at declining exchange rates: co-citation mining after
a probe of 60 resolved missed gold (``0/60 mention it in the
title\ldots{} text search structurally cannot reach this gold'');
submission padding derived from the scoring arithmetic (``with
$F1=2H/(N+G)$, an extra candidate pays whenever its hit probability
exceeds roughly $F1/2$''). &
A third model family for decorrelated errors (``when GPT and Claude
share a blind spot\ldots{} Gemini frequently doesn't''), with
dtype-visible execution diagnostics feeding an always-on judge. \\
\bottomrule
\end{tabular}
\caption{What each waypoint step implemented, tracked for one run per
task, with the producing session's stated transfer reasoning (quoted
verbatim from the sessions' reasoning records; one quote is a
winner-code comment). Steps are labeled by the shared waypoint grid;
per-arm magnitudes for all four arms are in Table~\ref{tab:waypoints}.}
\label{tab:mechanisms}
\end{table}

\input{appendix_inputs}

\section{Evolved-agent architecture sketches}
\label{app:arch}

One sketch follows for each agent featured in the main text's
What-evolution-built discussions
(Sections~\ref{sec:ds1000}--\ref{sec:pfb}): for each task, the board
leader, the cheapest point on the Pareto frontier, and one mid-price
agent. All are single-file Python
programs whose only LLM access is the menu of Appendix~\ref{app:menu};
sizes are total lines including comments and docstrings (evolution
documents its own designs).

\paragraph{\dsk{} 0.873 @ \$0.054 (1{,}226 lines~\agentlink{asta_ds1000}{v0_1_0_cap_0_127_fable51}{iter6_ensemble_exec_vote_escalate}; board leader).}
Three strong models from the three providers (Claude-Opus-4.8,
GPT-5.4 at medium reasoning, Gemini-3.1-Pro) write candidate solutions
in parallel. Each candidate runs in a fresh sandbox subprocess that
dumps its target variables with type, shape, dtype, and index; a
crashing candidate gets one repair round from Claude-Sonnet-4.6. When
the executed outputs agree, the majority wins outright with no judge
call, loop-free code preferred---\dsk{}'s hidden style rules sometimes
reject a correct value computed with an explicit loop, so a looping
winner may be rewritten, with the rewrite accepted only when sandbox
execution proves it returns the same value. A three-way split brings
in two more voices (Claude-Sonnet-4.6 and GPT-5.5; three of five
wins), and failing that a low-reasoning Opus judge picks among one
representative per distinct output. Agreement on the example is not
trusted blindly: when the majority's code differs, Sonnet writes an
alternative input and re-runs the agreeing candidates on it, sending
any divergence to the judge. The agent meters its own token spend,
caps optional steps, and always emits its best answer so far. It
converged on the core design of the earlier board leader (0.862 at
\$0.127, evolved by Opus-4.7 months before): multi-family candidates
executed in the sandbox, a judge on disagreement, an output-preserving
loop rewrite. That agent always ran four strong models plus Opus
critic calls; this one lets agreement skip the judge and escalates only
on disagreement, at 43\% of the cost.

\paragraph{\dsk{} 0.853 @ \$0.037 (500 lines~\agentlink{asta_ds1000}{v0_0_4_cap_0_08_opus}{iter11_ds1000_tridtype_judge}).}
This agent asks three medium-cost
models from the three providers for solutions at baseline effort
settings, executes all three in the free sandbox, and compares the
results at the level the grader actually checks: not just printed
values but the column-by-column dtypes of DataFrames and the dtypes of
arrays and Series, where \dsk{}'s hidden tests are strict. When every
candidate that ran cleanly agrees on that dtype-rich summary, the
agent submits immediately; on any disagreement it hands the
candidates, their outputs, and the diagnostics together to a single
high-effort adjudication call, then verifies and, if needed, repairs
the chosen answer.

\paragraph{\dsk{} 0.737 @ \$0.002 (302 lines~\agentlink{asta_ds1000}{v0_0_7_cap_0_003_opus}{iter12_strong_toplevel}; cheapest frontier point).}
The cheapest frontier agent is fundamentally the simplest program in
the family: one call to a relatively strong model behind a
deliberately short prompt. Against its \$0.003 budget, evolution's
chosen trade is a stronger model with a minimal 59-word preamble
rather than a cheaper model wrapped in machinery---a single GPT-5.4
call at the no-extended-reasoning default, whose short \dsk{} prompts
cost \$0.0013--0.0016 per problem, comfortably inside the free zone.
How it got there is the interesting part: early iterations built the
complex designs that win at higher price points---execution-verified
retries (iteration 2), an executed self-consistency ensemble (3),
reasoning escalation (4)---and the evolutionary meta-agent rejected
each based on experimental data. Its \texttt{reasoning.md} files note
that wrong \dsk{} answers execute cleanly, so crash-verification
catches nothing, and the marginal vote-or-verify call trades
correct-simple answers for over-clever broken ones. Prescriptive
prompting fared no better: a five-rule guide dropped the then
mini-based agent to 40\%, and a stricter contract preamble caused
indentation failures; only the tiny preamble survived. 

\paragraph{\pfb{} 0.501 @ \$0.232 (1{,}811 lines~\agentlink{asta_paper_finder}{v0_1_1_cap_0_279_fable52}{iter20_year_steered_expansion}; board leader).}
The board leader is the only \pfb{} entry of ours to draw on all three
providers. It routes each query by type, and for every type a
Claude-Opus-4.8 planner first turns the request into a plan. For
semantic queries, which are 73\% of the test set, the plan's keyword
and body-text searches run in parallel and their results are fused.
The benchmark's relevance criteria are hidden at query time, so the
planner infers its own and weights them; GPT-5.4-mini then rates every
candidate against each criterion, and a second tranche is graded only
when the first proves rich. Its signature move is a citation snowball:
starting from the papers that satisfy every criterion, the agent grades
their references and citers, taking first the papers linked to several
seeds. GPT-5.4 re-judges the strongest candidates before submission.
For named-paper queries, GPT-5.5 and Gemini-3.1-Pro give independent
readings of what the request refers to and Opus picks from the pooled
candidates. For metadata queries, the planner writes a structured
specification that deterministic filters apply, using citation counts
back-dated to the benchmark's cutoff. Spend is spread rather than
concentrated: GPT-5.4-mini 36\%, GPT-5.4 35\%, Opus 23\%.

\paragraph{\pfb{} 0.375 @ \$0.053 (2{,}097 lines~\agentlink{asta_paper_finder}{v0_0_7_cap_0_06_fable}{iter12_body_conjunction}).}
This agent's signature mechanism attacks queries whose answer lives in
the \emph{body} of papers rather than their titles or abstracts.  For example, the phrase ``rejection
sampling used in finetuning'' names a connection that keyword search
over titles never surfaces. A GPT-5.4 planner decomposes each query
into conjunctive constraints and issues full-text snippet searches
phrased the way a methods section would state the connection; each
such query feeds the candidate pool through its own retrieval channel
so body-matched papers get a real share of the pool. Candidates from
all channels are then graded in bulk by GPT-5.4-mini, with a selective
GPT-5.4 re-grade of the top of the ranking.

\paragraph{\pfb{} 0.220 @ \$0.006 (1{,}006 lines~\agentlink{asta_paper_finder}{v0_0_8_cap_0_033_opus}{iter9_rerank_rich_v1}; cheapest frontier point).}
The cheapest frontier agent never calls anything but GPT-5.4-mini, and
calls it at most twice per query. What replaces model strength is
breadth of search, since tool calls are free. On a semantic query, one
call rewrites the request as several keyword phrasings and names the
aspects a relevant paper must satisfy; searches on each of these build
the candidate pool; and a second call rates the top 90 candidates on
their abstracts, reserving the top rating for papers that satisfy every
aspect. Specific-paper and metadata queries follow the same pattern: one
call to interpret the request, tool calls to gather candidates, and at
most one call to choose among them. The result matches
ReAct/GPT-5-Mini on score at a tenth of that entry's spend (\$0.006
vs.\ \$0.060).

%% file: tables_slope.tex
\begin{table}[h]
\centering
\small
\begin{tabular}{llllrrc}
\toprule
Evol.\ model & Run & Task & $\tau$ (\$) & $\lambda/\tau$ & Util. & Table~\ref{tab:runs} \\
\midrule
Opus-4.7 & \texttt{paper\_finder-018} & \pfb{} & 0.355 & 0.10 & 2.5\% &  \\
 & \texttt{ds1000-049} & \dsk{} & 0.028 & 0.10 & 38.2\% &  \\
 & \texttt{ds1000-048} & \dsk{} & 0.028 & 0.10 & 50.7\% &  \\
Opus-4.8 & \texttt{paper\_finder-006} & \pfb{} & 0.033 & 0.09 & 18.0\% & $\checkmark$ \\
 & \texttt{paper\_finder-019} & \pfb{} & 0.355 & 0.10 & 49.0\% &  \\
 & \texttt{ds1000-044} & \dsk{} & 0.003 & 0.10 & 55.7\% & $\checkmark$ \\
 & \texttt{ds1000-050} & \dsk{} & 0.028 & 0.10 & 61.1\% &  \\
Opus-5 & \texttt{paper\_finder-011} & \pfb{} & 0.355 & 0.10 & 70.6\% & $\checkmark$ \\
 & \texttt{ds1000-045} & \dsk{} & 0.028 & 0.10 & 71.8\% &  \\
 & \texttt{paper\_finder-017} & \pfb{} & 0.063 & 0.10 & 81.0\% &  \\
 & \texttt{paper\_finder-009} & \pfb{} & 0.18 & 0.10 & 81.7\% &  \\
 & \texttt{paper\_finder-010} & \pfb{} & 0.063 & 0.10 & 83.2\% & $\checkmark$ \\
Fable-5 & \texttt{paper\_finder-012} & \pfb{} & 0.355 & 0.10 & 78.5\% & $\checkmark$ \\
 & \texttt{paper\_finder-016} & \pfb{} & 0.063 & 0.10 & 79.4\% &  \\
 & \texttt{paper\_finder-015} & \pfb{} & 0.063 & 0.10 & 81.0\% &  \\
 & \texttt{ds1000-047} & \dsk{} & 0.028 & 0.10 & 88.9\% &  \\
 & \texttt{paper\_finder-013} & \pfb{} & 0.063 & 0.10 & 92.2\% & $\checkmark$ \\
 & \texttt{ds1000-046} & \dsk{} & 0.028 & 0.10 & 98.6\% &  \\
Fable-5.1 & \texttt{ds1000-054} & \dsk{} & 0.127 & 0.10 & 42.8\% & $\checkmark$ \\
 & \texttt{paper\_finder-021} & \pfb{} & 0.063 & 0.10 & 92.4\% & $\checkmark$ \\
Fable-5.2 & \texttt{paper\_finder-024} & \pfb{} & 0.279 & 0.10 & 72.8\% &  \\
 & \texttt{paper\_finder-025} & \pfb{} & 0.279 & 0.10 & 73.5\% &  \\
 & \texttt{paper\_finder-023} & \pfb{} & 0.279 & 0.10 & 83.3\% & $\checkmark$ \\
\bottomrule
\end{tabular}
\caption{The 23 runs at $\lambda \le 0.1\tau$ behind Section~\ref{sec:manage}, item~3, grouped by evolution model. Util.\ is achieved mean agent cost over $\tau$ (official basis for board entries, internal basis otherwise). Runs marked in the last column are board entries and also appear in Table~\ref{tab:runs}; the rest are unsubmitted experiment runs. No run bought through its cap. Per-model ranges: Opus-4.7 2.5--51\% ($n=3$); Opus-4.8 18--61\% ($n=4$); Opus-5 71--83\% ($n=5$); Fable-5 78--99\% ($n=6$); Fable-5.1 43--92\% ($n=2$); Fable-5.2 73--83\% ($n=3$).}
\label{tab:slope}
\end{table}

%% file: appendix_inputs.tex
\section{Evolution background and objective documents, excerpted verbatim}
\label{app:inputs}

Below are the authored objective and
background documents of Section~\ref{sec:system} with each run's
$\tau$, $\lambda$, and environment notes substituted in: the
\pfb{} run at $\tau=\$0.355$ (the 0.432 board entry) and the
\dsk{} run at $\tau=\$0.003$ (the 0.737 board entry).
Objectives are reproduced in full; backgrounds are excerpted.
All non-bracketed text is word-for-word verbatim from the
documents, with their markdown rendered typographically;
bracketed italic lines mark skips and editorial notes.

\subsection{\pfb{} input (verbatim input for run at $\tau=\$0.355$)}

{\small
\subsubsection*{Domain Background}

\subsubsection*{PaperFindingBench (AstaBench)}

Each example is a literature-search query: a natural-language description of papers the user wants ("the BART paper", "papers by David Harel in Nature", "clustering-based attention in Transformers"), and a hidden gold relevance judgment. The agent returns a ranked list of Semantic Scholar \texttt{corpus\_id}s; each query gets an F1 score in [0, 1] (standard F1 by exact match for specific/metadata queries, LLM-judged adjusted F1 for semantic --- see the table below), and the overall score is the plain mean over queries.

\par}

{\small
\medskip\noindent\textbf{Three query types (\texttt{state.metadata["score\_type"]})}\par\nopagebreak

{\scriptsize\begin{center}\begin{tabular}{>{\raggedright\arraybackslash}p{0.62in}>{\raggedright\arraybackslash}p{0.85in}>{\raggedright\arraybackslash}p{1.55in}>{\raggedright\arraybackslash}p{1.05in}>{\raggedright\arraybackslash}p{0.35in}}
\toprule
score\_type & meaning & gold form & scoring path & train count \\
\midrule
\texttt{specific\_f1} & "the X paper" --- known target & \texttt{\{"corpus\_ids":[...]\}} & exact-match against \texttt{corpus\_ids} & 10 \\
\texttt{metadata\_f1} & author/year/venue filters & \texttt{\{"corpus\_ids":[...]\}} & exact-match against \texttt{corpus\_ids} & 8 \\
\texttt{semantic\_f1} & broad topical query & \texttt{\{"known\_to\_be\_good":[...], "known\_to\_be\_bad":[...], "relevance\_criteria":[\{name, description, weight\}, ...]\}} & LLM judge over each predicted paper, weighted by \texttt{relevance\_criteria} & 48 \\
\bottomrule
\end{tabular}\end{center}}

The held-out test set has a similar query-type mix, so improvements weighted by these proportions generalize.

All three paths produce a real-valued score in [0, 1]; differences in difficulty are large.

The query text is in \texttt{state.metadata["raw\_query"]}. The full \texttt{state.input} wraps it in a longer instruction template; either is fair game.

\par}

\medskip\noindent\emph{[\dots\ the output-schema section (JSON shape and field
conventions), except the grounding requirement:]}\medskip

{\small
\begin{itemize}\setlength{\itemsep}{1pt}
\item \textbf{Grounding requirement}: \texttt{markdown\_evidence} must be up to 8 passages quoted \textbf{verbatim} from text you retrieved for that same paper (title/abstract/tldr/snippet returned by the tools), joined by \texttt{ ... }. Each passage is checked independently: any passage not verbatim-derivable from retrieved corpus text is discarded before the judge sees it, and the judge scores the paper on whatever grounded passages remain. If \emph{every} passage is discarded, the paper is scored Not Relevant with no judge call. Punctuation/case/whitespace differences are tolerated; paraphrased or invented passages are not, and earn nothing. Evidence beyond 2500 characters per paper is truncated before the grounding check and the judge --- text past the cap earns nothing (clipped papers are listed per-problem in \texttt{evidence\_truncation.md}).
\end{itemize}
\par}

{\small
\medskip\noindent\textbf{Available tools (\texttt{state.tools})}\par\nopagebreak

The PaperFindingBench task attaches the \textbf{Asta MCP corpus tools} --- eight of them. A date-cutoff filter is applied task-side so results don't leak papers published after the benchmark snapshot --- with one gap: the citing-paper lists returned by \texttt{get\_citations} are not filtered (see the search-semantics notes below).

\par}

\medskip\noindent\emph{[\dots\ the eight-tool parameter table and return-shape parsing
helpers]}\medskip

{\small
\medskip\noindent\textbf{Search semantics (verified against the live server)}\par\nopagebreak

\begin{itemize}\setlength{\itemsep}{1pt}
\item \textbf{Term matching is lenient, with no query operators.} Extra or missing terms don't zero a result set (adding a gibberish term leaves the top hits unchanged); quoting a phrase does NOT enforce its presence (a quoted nonexistent phrase still returns full results); \texttt{-term} does not exclude; \texttt{OR} is treated as an ordinary token. Query text steers \emph{ranking} only.
\item \textbf{Interrogative/imperative framing returns ZERO hits.} "Could you suggest research that investigates X?" $\rightarrow$ 0 results, with or without punctuation; the bare noun phrase "X" $\rightarrow$ full results, even with articles and prepositions intact. Strip the question/request preamble; keyword or noun-phrase queries only. (\texttt{snippet\_search} is tolerant of full natural-language queries --- it's the right tool for sentence-shaped input.)
\end{itemize}
\par}

{\small
\begin{itemize}\setlength{\itemsep}{1pt}
\item The docstring prose on some tools states stale defaults ("fields default is title", "limit default is 50") inherited from the upstream server --- \textbf{trust the parameter defaults instead} (a rich field set including \texttt{corpusId}, limit 20). Also, the docstrings' "available fields" lists omit \texttt{corpusId} even though it's valid and essential --- when trimming \texttt{fields}, always keep \texttt{title,abstract,corpusId}.
\end{itemize}
\par}

\medskip\noindent\emph{[\dots\ six further verified-semantics notes and the
transport/timeout/rate-limit layer]}\medskip

\medskip\noindent\emph{[\dots\ the LLM-calls section: the nine-handle price table
(Table~\ref{tab:menu}) and per-call conventions]}\medskip

{\small
\medskip\noindent\textbf{Standard Tools constraint}\par\nopagebreak

This benchmark targets the \textbf{Standard Tools} leaderboard tier. The agent may use only:

\begin{itemize}\setlength{\itemsep}{1pt}
\item Tools attached to \texttt{state.tools} (the Asta MCP corpus suite)
\item LLM calls through \texttt{model\_registry} handles (Inspect-tracked)
\item Standard Python (json, re, asyncio, dataclasses, ...)
\end{itemize}

It must \textbf{not} import third-party search backends (Elasticsearch, Pinecone, custom indices), nor the AI2-internal Mabool client (\texttt{paper\_finder\_ai2i}), nor call web APIs directly --- including the public Semantic Scholar API (\texttt{api.semanticscholar.org}): the \texttt{state.tools} suite is the agent's only corpus access. Those tools enforce the benchmark's snapshot date-cutoff; the live public API does not, so calling it both breaks the Standard Tools tier and leaks post-snapshot papers the scorer treats as wrong. It must also not persist retrieval results across queries --- every evaluation's papers must come through the tools (within-query in-memory bookkeeping over tool results is normal and fine). The evaluator may reject candidates that import outside an allowlist.

\par}

{\small
\medskip\noindent\textbf{Per-query cost}\par\nopagebreak

Your cost is the LLM calls your agent makes through \texttt{model\_registry} handles, recorded per problem as \texttt{eval\_cost} in \texttt{result.json}. Tool calls (\texttt{paper\_search}, \texttt{snippet\_search}, the rest of the MCP suite) are free, in unlimited quantity.

\par}

\medskip\noindent\emph{[\dots\ the relevance-judge section: the judge grades each
returned paper from its \texttt{markdown\_evidence}
alone against the query's weighted criteria;
per-query diagnostics expose the criteria, verdicts,
and exact score arithmetic post-hoc; judge spend is
recorded but never penalized]}\medskip

{\small
\medskip\noindent\textbf{Iteration-aggregate score}\par\nopagebreak

Per-example scoring is continuous F1 in [0, 1]. At the end of each iteration, your batch is combined into a single score: your mean F1 (on a 0--100 scale) minus a cost penalty when your mean batch agent-spend exceeds the threshold. The penalty is expressed in fully-wrong-query units --- each \$0.0355 of mean spend over \$0.355 subtracts one error-equivalent (one query's worth of F1) from your score. Only \texttt{model\_registry} handle calls are metered --- tool calls are free.

\par}

\begin{table}[h]
\centering
\small
\begin{tabular}{lp{3.6in}}
\toprule
Mean agent cost & Effect on score \\
\midrule
$\le$ \$0.355 & No effect on score --- two free-zone agents with the same raw mean F1 score identically, regardless of their actual spend \\
\$0.355--\$0.3905 & Tiebreaker --- lose tied F1 to a cheaper agent; \textbf{need 1+ more fully-correct query} to win \\
\$0.3905--\$0.426 & \textbf{Need 2+ more fully-correct queries} than a free-zone agent to win \\
\$0.426--\$0.4615 & \textbf{Need 3+ more fully-correct queries} than a free-zone agent to win (in practice, a decisive penalty) \\
\dots{} & Each additional \$0.0355 of mean spend adds 1 to the breakeven count \\
\bottomrule
\end{tabular}
\caption{\emph{The cost-penalty table exactly as substituted into this
run's background document ($\tau=\$0.355$, $\lambda=\$0.0355$);
each run sees its own instantiation. This is
Eq.~\ref{eq:objective} rendered for the optimizer as breakeven
arithmetic.}}
\label{tab:penalty}
\end{table}

{\small
\medskip\noindent\textbf{Time budget}\par\nopagebreak

Your agent times out and the query scores 0 if a single query takes more than \textbf{29 minutes} of wall-clock. Per-query wall-clock is recorded as \texttt{eval\_wall\_clock\_seconds} in each problem's \texttt{result.json}.

\par}

\medskip\noindent\emph{[\dots\ the exact per-query scoring formulas (described in
Appendix~\ref{app:tasks})]}\medskip

{\small
\medskip\noindent\textbf{Diagnostics}\par\nopagebreak

Any \texttt{print()} output from the agent is captured and included in evaluation diagnostics as \texttt{agent\_stdout}. Use \texttt{print()} to log anything you think would be helpful for you to see when improving the agent in later rounds.

\textbf{Session-side corpus access (yours, not the agent's).} Two read-only surfaces are available to you for analysis; neither may appear in agent code --- the agent's only corpus access is \texttt{state.tools} (see the Standard Tools constraint).

\begin{itemize}\setlength{\itemsep}{1pt}
\item \textbf{The tool probe} --- \texttt{python ../../session\_tools/tool\_probe.py} from your workspace --- calls the same Asta MCP corpus tools through the same task-side wrappers the evaluation applies (same snapshot date-cutoff, same field defaults, same retry behavior), so its output is exactly what your agent's own tool call would return. \texttt{--list} prints the eight tools and their parameters; arguments are \texttt{key=value} pairs, e.g. \texttt{python ../../session\_tools/tool\_probe.py search\_papers\_by\_relevance keyword="sparse attention" limit=5}.
\item \textbf{The public Semantic Scholar API} (\texttt{api.semanticscholar.org}) --- the live world-view, useful precisely where the probe's snapshot view is not: resolving gold \texttt{corpus\_id}s to titles and publication dates, or checking whether an id postdates the snapshot. Batch every id into one call (\texttt{POST /graph/v1/paper/batch} with ids like \texttt{CorpusId:123}); unauthenticated per-id calls hit the rate limit. Its records are live and can differ from what the tools return.
\end{itemize}
\par}

{\small
\subsubsection*{Domain Objective}

Evolve a PaperFindingBench agent that, given a natural-language literature-search query, returns a list of Semantic Scholar corpus\_ids maximizing adjusted micro-F1 against the query's hidden gold, using only the Standard tools (Asta MCP corpus + model\_registry LLM handles).

Your primary goal is simple: maximize the score on held-out queries. The scoring function (described in Domain Background in CLAUDE.md) encodes this directly --- retrieval quality is the dominant signal, and cost acts as a tiebreaker close to threshold but starts to actively trade off against score farther out. Cost means your LLM spend through \texttt{model\_registry} handles (tool calls are free), and the free zone is the batch \emph{average}, not per-query: you can spend more on hard queries and less on easy ones. Above \$0.355 per query on average, every \$0.0355 of extra spend costs you one fully-wrong-query-equivalent of score; see the cost-penalty table in Domain Background for the breakeven math.

Each iteration draws a different sample of queries, and the final agent is evaluated on a held-out test set it has never seen. So, to rephrase your goal, your objective is to build an agent that generalizes to unseen queries --- the visible batch is a training signal, not the target.

\par}

{\small
\subsubsection*{Evolution Environment}

\medskip\noindent\textbf{Available Data}\par\nopagebreak

Use your available tools to explore the experiment directory. Key artifacts:

\begin{itemize}\setlength{\itemsep}{1pt}
\item \texttt{../../agents/<name>/} --- agent source code (one directory per agent)
\item \texttt{../../iteration\_NNN/} --- evaluation results per iteration:
\item \texttt{error\_analysis\_report.md} --- cross-agent score comparison and failure summary
\item \texttt{error\_index.json} --- machine-readable score data (error analysis report was derived from this)
\item \texttt{cost\_report.md} --- per-agent cost breakdown. Useful if you are instructed to pay attention to cost
\item \texttt{agent\_<name>/problems/<id>/} --- per-problem results and diagnostics
\end{itemize}

During testing and refinement rounds (Rounds 2+), your results appear in \texttt{./iteration\_NNN\_test/} (in your working directory, not the experiment root).

\par}

\medskip\noindent\emph{[\dots\ strategy-tools, scratch-space, and CLI notes]}\medskip

\subsection{\dsk{} input (verbatim input for run at $\tau=\$0.003$)}

{\small
\subsubsection*{Domain Background}

\subsubsection*{DS-1000 (AstaBench)}

Each example is a Python data-science problem. The agent receives a natural-language question with an embedded code skeleton and must emit Python code that, when appended to the program, makes a variable called \texttt{result} hold the correct value.

\par}

\medskip\noindent\emph{[\dots\ an illustrative example problem and the solver-state
field table (a real pool problem is reproduced in
Appendix~\ref{app:examples})]}\medskip

{\small
\medskip\noindent\textbf{Required output}\par\nopagebreak

Write a single \texttt{<code>...</code>} block to \texttt{state.output.completion}. The opening \texttt{<code>} and closing \texttt{</code>} tags \textbf{are required} --- the scorer uses them to extract the answer. Everything between them is appended to a hidden test program that exercises \texttt{result} against test inputs.

{\footnotesize\begin{verbatim}
<code>
result = a[a != 0]
</code>
\end{verbatim}}

Inside the tags: executable Python only. No prose, no markdown fences (\texttt{ }\texttt{}python \texttt{), no }BEGIN SOLUTION\texttt{ / }END SOLUTION\texttt{ markers. Python }\#` comments are optional.

Outside the tags (i.e., in \texttt{state.output.completion} before \texttt{<code>} or after \texttt{</code>}): nothing. Don't preface the answer with a chain-of-thought summary --- the scorer doesn't see it and it just adds tokens.

\par}

{\small
\medskip\noindent\textbf{The Docker sandbox}\par\nopagebreak

\texttt{python\_session} runs Python inside a Docker container with a curated data-science package set: pandas, numpy, scipy, scikit-learn, statsmodels, matplotlib, seaborn, gensim, torch, tensorflow-cpu, xgboost (versions pinned to AstaBench's compose). Each sample gets a fresh container; variables persist within a sample across multiple \texttt{python\_session} calls (Jupyter-kernel-like). Default cell timeout: 5 minutes. Working directory: \texttt{/workspace/}.

\par}

\medskip\noindent\emph{[\dots\ the API-surface examples and the LLM-calls section
(price table: Table~\ref{tab:menu})]}\medskip

{\small
\medskip\noindent\textbf{Scoring}\par\nopagebreak

The agent's \texttt{<code>} block is extracted, concatenated with hidden setup and test code, and run inside the sandbox. The score for the sample is \textbf{1.0} if the appended code makes \texttt{result} match the reference under all hidden test inputs, else \textbf{0.0}. No partial credit.

A subset of problems additionally enforce \textbf{style/idiom constraints on the submitted code itself}. Two flavors appear: (1) forbidding Python control-flow constructs like \texttt{for}/\texttt{while} to push toward library calls, and (2) requiring a specific library function name to appear in the solution, ruling out manual reimplementations. The constraint is sometimes flagged in the prompt ("without using X", "the efficient way", "not one by one") but is more often implicit in the spirit of the question: asking \emph{"how do I do X with NumPy"} invites a NumPy-idiomatic answer, and a workaround that bypasses the library can fail even when the output is correct. When this happens, the per-problem \texttt{test\_result.md} shows an assertion raised from a \texttt{test\_string} function (versus correctness failures, which raise from \texttt{test\_execution}). Both outcomes score 0.0; the traceback tells the agent whether to fix the \emph{answer} or the \emph{form}.

\par}

{\small
\medskip\noindent\textbf{Iteration-aggregate score}\par\nopagebreak

Per-example scoring is binary correctness (1.0 or 0.0). At the end of each iteration, your batch is combined into a single score: your accuracy (on a 0--100 scale) minus a cost penalty when your mean batch spend exceeds the threshold. The penalty is expressed in wrong-answer units --- each \$0.0003 of mean spend over \$0.003 subtracts one error-equivalent from your score. Only \texttt{get\_model()} calls are metered --- \texttt{python\_session} and \texttt{sandbox()} don't count.

\par}

\medskip\noindent\emph{[\dots\ the cost-penalty table, analogous to
Table~\ref{tab:penalty} at $\tau=\$0.003$,
$\lambda=\$0.0003$]}\medskip

{\small
\medskip\noindent\textbf{Time budget}\par\nopagebreak

Your agent times out and the problem scores 0 if a single problem takes more than \textbf{29 minutes} of wall-clock. This is a generous budget and is unlikely to be the binding constraint. Per-problem wall-clock is recorded as \texttt{eval\_wall\_clock\_seconds} in each problem's \texttt{result.json}.

\par}

\medskip\noindent\emph{[\dots\ diagnostics notes]}\medskip

{\small
\subsubsection*{Domain Objective}

Evolve a DS-1000 agent that, given a Python data-science problem prompt, produces a \texttt{<code>...</code>} block whose contents make the hidden test program's \texttt{result} variable match the reference under all hidden test inputs.

Your primary goal is simple: maximize the score on held-out problems. The scoring function (described in Domain Background in CLAUDE.md) encodes this directly --- correctness is the dominant signal, and cost acts as a tiebreaker close to threshold but starts to actively trade off against correctness farther out. The free zone is the batch \emph{average}, not per-problem: you can spend more on some problems and less on others. Above \$0.003 per problem on average, every \$0.0003 of extra spend costs you one error-equivalent of score; see the cost-penalty table in Domain Background for the breakeven math.

Each iteration draws a different sample of problems, and the final agent is evaluated on a held-out test set it has never seen. After you construct your agent, it will be tested on entirely new batches of examples in future iterations. So, to rephrase your goal, your objective is to build an agent that generalizes to unseen problems --- the visible batch is a training signal, not the target.

\par}

\medskip\noindent\emph{[\dots\ the Evolution Environment section, identical to the
\pfb{} one above]}\medskip

\subsection{Seed agents}

The \dsk{} seed (46 lines), verbatim:

{\footnotesize\begin{verbatim}
"""Baseline DS-1000 solver.

One-shot baseline: send the problem prompt to the LLM, take the response,
and emit the required `<code>...</code>` block.
"""

from inspect_ai.model import GenerateConfig
from inspect_ai.solver import Generate, TaskState, solver

# LLM handles are imported from `model_registry`. Pick one per call,
# or mix across calls. See CLAUDE.md (Domain Background) for the
# full list of handles and their pricing.
from model_registry import GPT_5_4_MINI

def _wrap_in_code_tags(text: str) -> str:
    """Ensure the response is wrapped in `<code>...</code>` tags.

    The DS-1000 scorer's `postprocess` strips a few common envelopes
    (```python fences, <code> tags, END SOLUTION markers). This helper
    is conservative: if the model already produced `<code>...</code>`
    or a markdown fence, leave it alone; otherwise wrap raw code.
    """
    s = text.strip()
    if "<code>" in s and "</code>" in s:
        return s
    if s.startswith("```"):
        return s
    return f"<code>\n{s}\n</code>"

@solver
def make_solver():
    async def solve(state: TaskState, generate: Generate) -> TaskState:
        # Demonstration print statement -- captured in `agent_stdout`
        # alongside the per-problem diagnostics, so anything you print
        # here is available for retrospective analysis.
        print(f"[{state.sample_id}] library={state.metadata.get('library', '?')}")

        resp = await GPT_5_4_MINI.generate(state.input)
        completion = resp.completion or ""
        state.output.completion = _wrap_in_code_tags(completion)
        print(f"  emitted {len(state.output.completion)} chars")
        return state

    return solve
\end{verbatim}}

The \pfb{} seed (121 lines) follows the same conventions;
\emph{[paraphrased as pseudocode]}:

{\footnotesize\begin{verbatim}
read query from state.metadata
keywords <- GPT_5_4_MINI("extract a concise keyword search query
                          (3-8 words) from this request: " + query)
  # load-bearing: the keyword search returns ZERO hits for full
  # natural-language questions; guard against empty completions
hits <- search_papers_by_relevance(keyword=keywords,
            fields="title,abstract,corpusId", limit=30)
indices <- GPT_5_4_MINI("given the query and these candidate
                         titles, return up to 10 relevant indices")
kept <- hits[indices], padded best-first to >= 8 papers
  # near-empty semantic lists degenerate the rank term
submit JSON {query_id, results: [{paper_id: str(corpusId),
        markdown_evidence: title + " -- " + abstract[:400]}]}
\end{verbatim}}